\documentclass[journal]{IEEEtran}
\usepackage{ifpdf}
\usepackage{booktabs,amsfonts}
\usepackage{color}
\usepackage{array}
\usepackage{multirow}

\newcommand{\etal}{\textit{et al}.}
\newcommand{\ie}{\textit{i}.\textit{e}., }
\newcommand{\eg}{\textit{e}.\textit{g}., }
\newcommand{\etc}{\textit{etc}}

\usepackage{cite}
\ifCLASSINFOpdf
   \usepackage[pdftex]{graphicx}
\else
\fi
\usepackage{amsmath}
\usepackage{algorithmic}
\usepackage{array}
\begin{document}
%
% paper title
% Titles are generally capitalized except for words such as a, an, and, as,
% at, but, by, for, in, nor, of, on, or, the, to and up, which are usually
% not capitalized unless they are the first or last word of the title.
% Linebreaks \\ can be used within to get better formatting as desired.
% Do not put math or special symbols in the title.
\title{SOLVER: Scene-Object Interrelated Visual\\Emotion Reasoning Network}
%
%
% author names and IEEE memberships
% note positions of commas and nonbreaking spaces ( ~ ) LaTeX will not break
% a structure at a ~ so this keeps an author's name from being broken across
% two lines.
% use \thanks{} to gain access to the first footnote area
% a separate \thanks must be used for each paragraph as LaTeX2e's \thanks
% was not built to handle multiple paragraphs
%

\author{Jingyuan~Yang,\IEEEmembership{}
	Xinbo~Gao,~\IEEEmembership{Senior~Member,~IEEE},
 	Leida~Li,~\IEEEmembership{Member,~IEEE},
	Xiumei~Wang,\IEEEmembership{}
	and~Jinshan~Ding\IEEEmembership{}% <-this % stops a space
	\vspace{-10pt}
	\thanks{
	Manuscript received February 10, 2021; revised July 7, 2021 and August 14, 2021; accepted September 21, 2021. 
	This work was supported in part by the National Natural Science Foundation of China under Grants 62036007, 62050175, 61772402 and 62171340.
	The associate editor coordinating the review of this manuscript and approving it for publication was Prof. Husrev Sencar. (\textit{Corresponding author: Xinbo Gao.})
	
	J. Yang, X. Wang and J. Ding are with the School of Electronic Engineering, Xidian University, Xi'an 710071, China (e-mail: jingyuanyang@stu.xidian.edu.cn; wangxm@xidian.edu.cn; ding@xidian.ed-u.cn).
		
	X. Gao is with the School of Electronic Engineering, Xidian University, Xi'an 710071, China (e-mail: xbgao@mail.xidian.edu.cn) and with
	the Chongqing Key Laboratory of Image Cognition, Chongqing University of Posts and Telecommunications, Chongqing 400065, China (e-mail:
	gaoxb@cqupt.edu.cn).
	
	L. Li is with the School of Artificial Intelligence, Xidian University, Xi'an 710071, China	(e-mail: ldli@xidian.edu.cn).
}%<-this % stops a space}
}

% note the % following the last \IEEEmembership and also \thanks - 
% these prevent an unwanted space from occurring between the last author name
% and the end of the author line. i.e., if you had this:
% 
% \author{....lastname \thanks{...} \thanks{...} }
%                     ^------------^------------^----Do not want these spaces!
%
% a space would be appended to the last name and could cause every name on that
% line to be shifted left slightly. This is one of those "LaTeX things". For
% instance, "\textbf{A} \textbf{B}" will typeset as "A B" not "AB". To get
% "AB" then you have to do: "\textbf{A}\textbf{B}"
% \thanks is no different in this regard, so shield the last } of each \thanks
% that ends a line with a % and do not let a space in before the next \thanks.
% Spaces after \IEEEmembership other than the last one are OK (and needed) as
% you are supposed to have spaces between the names. For what it is worth,
% this is a minor point as most people would not even notice if the said evil
% space somehow managed to creep in.

% The paper headers
\markboth{IEEE TRANSACTIONS ON IMAGE PROCESSING}%
{Shell \MakeLowercase{\textit{et al.}}: Bare Demo of IEEEtran.cls for IEEE Journals}
% The only time the second header will appear is for the odd numbered pages
% after the title page when using the twoside option.
% 
% *** Note that you probably will NOT want to include the author's ***
% *** name in the headers of peer review papers.                   ***
% You can use \ifCLASSOPTIONpeerreview for conditional compilation here if
% you desire.

% If you want to put a publisher's ID mark on the page you can do it like
% this:
%\IEEEpubid{0000--0000/00\$00.00~\copyright~2015 IEEE}
% Remember, if you use this you must call \IEEEpubidadjcol in the second
% column for its text to clear the IEEEpubid mark.

% use for special paper notices
%\IEEEspecialpapernotice{(Invited Paper)}

% make the title area
\maketitle

% As a general rule, do not put math, special symbols or citations
% in the abstract or keywords.
\begin{abstract}
Visual Emotion Analysis (VEA) aims at finding out how people feel emotionally towards different visual stimuli, which has attracted great attention recently with the prevalence of sharing images on social networks.
Since human emotion involves a highly complex and abstract cognitive process, it is difficult to infer visual emotions directly from holistic or regional features in affective images.
% a certain pixel or object
It has been demonstrated in psychology that visual emotions are evoked by the interactions between objects as well as the interactions between objects and scenes within an image.
Inspired by this, we propose a novel~\textit{Scene-Object interreLated Visual Emotion Reasoning network (SOLVER)} to predict emotions from images.
To mine the emotional relationships between distinct objects, we first build up an~\textit{Emotion Graph} based on semantic concepts and visual features.
Then, we conduct reasoning on the Emotion Graph using Graph Convolutional Network (GCN), yielding emotion-enhanced object features.
We also design a~\textit{Scene-Object Fusion Module} to integrate scenes and objects, which exploits scene features to guide the fusion process of object features with the proposed scene-based attention mechanism.
Extensive experiments and comparisons are conducted on eight public visual emotion datasets, and the results demonstrate that the proposed SOLVER consistently outperforms the state-of-the-art methods by a large margin.
Ablation studies verify the effectiveness of our method and visualizations prove its interpretability, which also bring new insight to explore the mysteries in VEA.
Notably, we further discuss SOLVER on three other potential datasets with extended experiments, where we validate the robustness of our method and notice some limitations of it.

\end{abstract}

% Note that keywords are not normally used for peerreview papers.

% For peer review papers, you can put extra information on the cover
% page as needed:
% \ifCLASSOPTIONpeerreview
% \begin{center} \bfseries EDICS Category: 3-BBND \end{center}
% \fi
%
% For peerreview papers, this IEEEtran command inserts a page break and
% creates the second title. It will be ignored for other modes.
\IEEEpeerreviewmaketitle

\section{Introduction}
\label{sec:introduction}
\begin{figure}
	\centering
	\includegraphics[width=\linewidth]{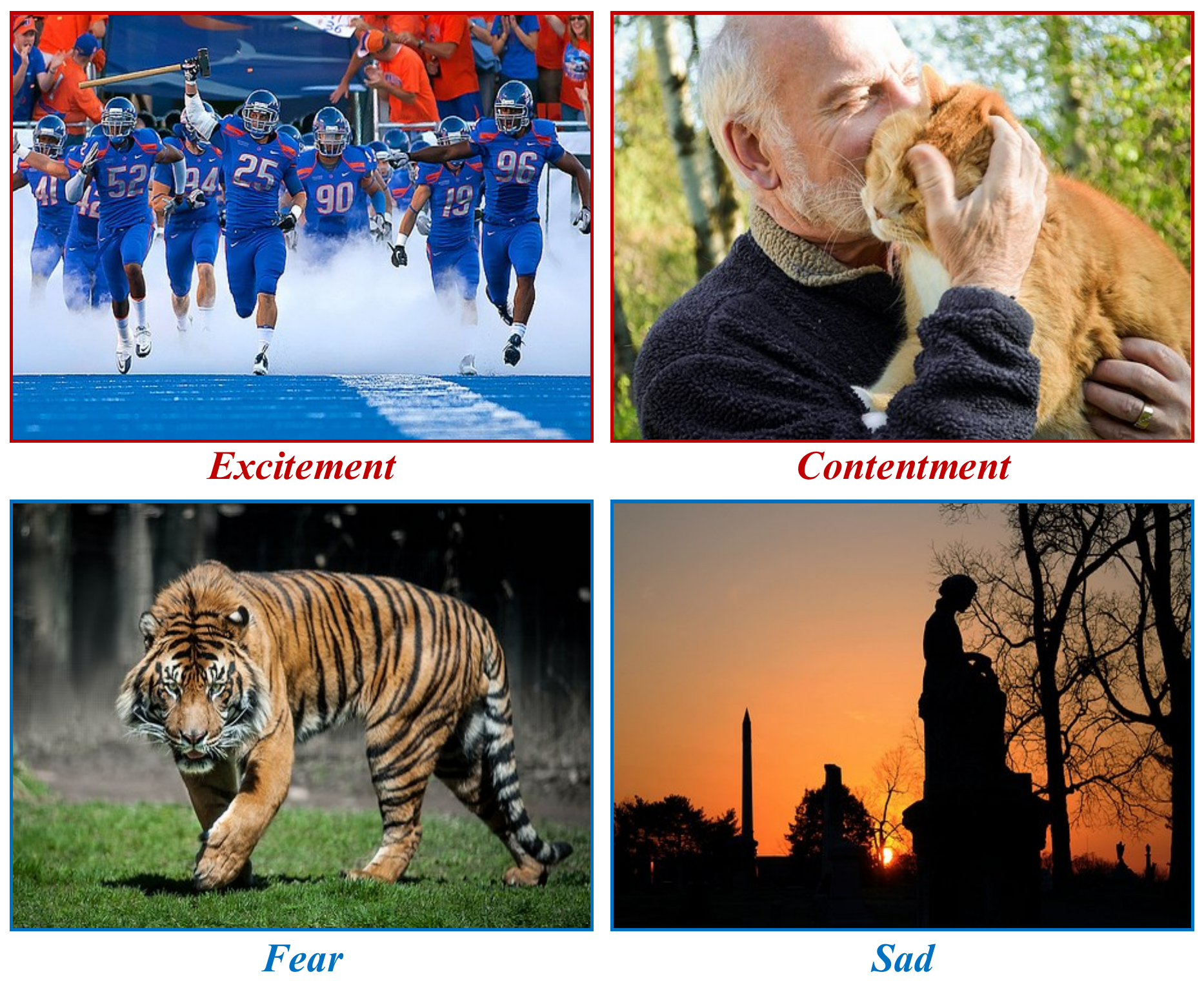}
	\vspace{-10pt}
	\caption{Four affective images with different emotions from FI dataset. These images deliver not only objects and scenes, but also, more significantly for us humans, emotions that are evoked involuntarily.}
	\vspace{-5pt}
	%	\Description{}
	\label{fig:intro_1}
\end{figure}

Human beings are born with one of the greatest power --- \textit{emotion}~\cite{gerber1985rage}, which is invisible yet indispensable in our daily lives.
With the prevalence of social networks, images have become a major medium for people to express emotions and to understand others as well. 
Nowadays, computer vision algorithms mostly teach networks how to ``see'' like a human while scarcely tell them how to ``feel'' like a human,
which has been considered as a crucial step towards the understanding of human cognition~\cite{phelps2006emotion}. 
%which is regarded as a crucial step towards strong artificial intelligence~\cite{de2013personal}.
As shown in Fig.~\ref{fig:intro_1}, when encountering an image, not only do we see objects and scenes, but we are also evoked by a certain emotion behind them involuntarily.
Therefore, aiming at finding out how people feel emotionally towards different visual stimuli, \textit{Visual Emotion Analysis (VEA)} has become an important research topic with increasing attention recently. 
Progress in VEA may benefit other related tasks (\eg image aesthetic assessment~\cite{li2020personality,zeng2019unified}, stylized image captioning~\cite{chen2018factual,guo2019mscap}, and social relation inference~\cite{zhang2015learning}), and will have a great impact on many applications, including decision making~\cite{szczepanowski2013perception}, smart advertising~\cite{mitchell1986effect}, opinion mining~\cite{li2019survey}, and mental disease treatment~\cite{wieser2012reduced}.
\begin{figure*}
	\centering
	\includegraphics[width=\textwidth]{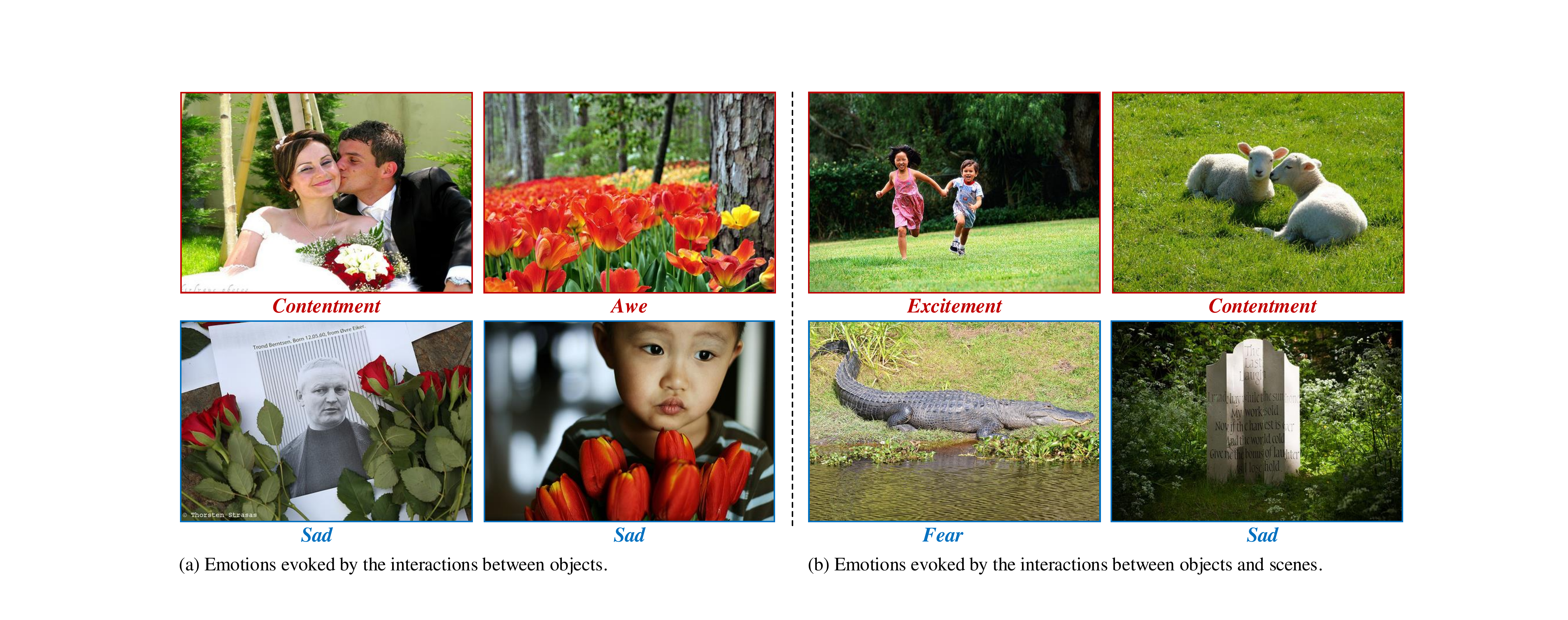}
	\vspace{-15pt}
	\caption{Examples from FI dataset. We believe that visual emotions are evoked by the interactions between objects (a) as well as the interactions between objects and scenes (b) within an image. In (a), \textit{red flowers} interact with different objects to convey distinct emotions. In (b), under the same scene of \textit{lawn}, diverse scene-object interactions evoke different emotions.}
	\vspace{-5pt}
	%	\Description{}
	\label{fig:intro_4}
\end{figure*}

Researchers have been engaged in VEA for more than two decades~\cite{lang1997international}, during which methods have varied from the early traditional ones to the recent deep learning ones. 
In the early years, researchers designed hand-crafted features (\eg color, texture, shape, composition, balance, and emphasis) based on art and psychological theories~\cite{machajdik2010affective,zhao2014exploring,siersdorfer2010analyzing,zhao2014affective,borth2013large}, attempting to find out the potential impact factors on human visual emotions.
However, it was hard to cover all the important factors by implementing manually designed features, which led to sub-optimal results. 
With the prevalence of deep learning networks, more and more researchers in VEA employed~\textit{Convolutional Neural Networks (CNNs)} in an end-to-end manner to predict emotions~\cite{chen2014deep,rao2016learning,yang2018visual,yang2018weakly,yang2018retrieving,yang2017joint,you2015robust,peng2016emotions}.
Earlier attempts directly implemented a general CNN to extract holistic features from affective images~\cite{chen2014deep,you2015robust}, which neglected the fact that visual emotions can also be evoked by local regions.
Recently, in order to trace emotions more concretely, researchers in VEA adopted detection methods and attention mechanisms to focus on local regions~\cite{peng2016emotions,yang2018visual,yang2018weakly}.
Most of the aforementioned methods mapped holistic or regional features to emotion labels directly. 
However, since human emotions involves a highly complex and abstract cognitive process, we argue that a direct mapping may underestimate the wide \textit{affective gap}~\cite{yang2018visual} between low-level pixels and high-level emotions. 

In addition to computer vision, researchers in other fields also devoted themselves to exploring the mysteries of visual emotions, including psychology~\cite{strongman1996psychology}, neuroscience~\cite{lane2002cognitive} and sociology~\cite{hochschild1998sociology}.
It has been demonstrated in psychology that scenes and objects can be regarded as emotional stimuli in affective images~\cite{brosch2010perception}.
In~\cite{frijda2009emotion}, psychologist Frijda suggested that emotional state is not about a specific object, but the perception of multiple emotionally meaningful objects.
Neuroscientist Moshe suggested that visual objects occur in rich surroundings are often embedded with other related objects, which serves as a key cognitive process in the human brain~\cite{bar2004visual}.

Based on the above studies, we believe that visual emotions are evoked by the interactions between objects as well as the interactions between objects and scenes within an image.
To be specific, we argue that rather than an isolated object, multiple objects in an image interact with each other and jointly contribute to the final emotion.
Besides, as scenes affect the emotional tone of an image, we further take scenes into account by mining the scene-object interactions, assuming that it is scenes that guide objects to evoke distinct emotions.
As shown in Fig.~\ref{fig:intro_4}(a), there are red roses in both left images and red tulips in the ones on the right, but we have different emotions towards them. 
Take red roses as an example, when red roses appear with a bride and a groom, it can be inferred as a wedding ceremony, which brings people a positive emotion, \ie \textit{contentment}. 
Oppositely, when red roses and a deadee comes together, people may feel \textit{sad} with a negative emotion.
Therefore, it is not the red roses alone that evoke a specific emotion, but the interactions between red roses and other objects jointly determine the final result. 
Fig.~\ref{fig:intro_4}(b) shows that under the same scene of lawn, different objects may evoke different emotions as well.
In particular, running kids on the lawn may bring us \textit{excitement} while a sly crocodile lying on the lawn makes us \textit{fear}.
It is obvious that different interactions between objects and scenes may evoke distinct emotions, from which we attach importance to both objects and scenes when analyzing visual emotions.

Motivated by the above facts, we propose a novel Scene-Object interreLated Visual Emotion Reasoning network (SOLVER), aiming at predicting visual emotions from the interactions between objects and objects as well as objects and scenes.
In order to mine the emotional relationships between different objects within an image, we construct an Emotion Graph and conduct reasoning on it.
To be specific, by transforming and filtering detected object features, we construct the Emotion Graph with objects as nodes and emotional relationships as edges.
Subsequently, we adopt~\textit{Graph Convolutional Network (GCN)} to perform reasoning on the Emotion Graph, which correlates objects with their emotional relationships and eventually yields emotion-enhanced object features.
Furthermore, we propose a Scene-Object Fusion Module to interrelate scenes and objects with each other.
Specifically, a novel scene-based attention mechanism is designed by exploiting scene features as guidance in object fusion process, which not only fuses multiple object features into a single one, but also serves as a pivot to interact between objects and scenes.

Our contributions can be summarized as follows:
\begin{itemize}
	\item We propose a novel framework, namely SOLVER, to predict visual emotions from the interactions between objects and objects as well as objects and scenes within an image, which outperforms the state-of-the-art methods on eight public visual emotion datasets.
	To the best of our knowledge, it is the first work that reasons between objects and scenes to infer emotions.
	\item We construct an Emotion Graph to depict emotional relationships between different objects, and subsequently conduct GCN reasoning on it, aiming to interrelate objects with their emotional relationships.
	\item We propose a Scene-Object Fusion Module by exploiting scene features to guide object fusion process with a designed scene-based attention mechanism, which serves as a pivot to mine the emotional relationships between objects and scenes.
\end{itemize}

The rest of the paper is organized as follows. Section~\ref{sec:related_work} overviews the existing methods on visual emotion analysis, object detection, and graph reasoning. In Section~\ref{sec:methodology}, we introduce the proposed SOLVER by constructing an Emotion Graph and a Scene-based Fusion Module. Extensive experiments, including comparisons, ablation studies, visualizations, and further discussions are conducted on visual emotion datasets and other potential datasets given in Section~\ref{sec:experimental_results}. Finally, we conclude our work in Section~\ref{sec:conclusion}.

\section{Related work}
\label{sec:related_work}
This work addresses the problem of visual emotion analysis, which is also closely related to object detection and graph reasoning.
In this section, we review the existing methods from the above three aspects.

\subsection{Visual Emotion Analysis}
\label{sec:visual_emotion_analysis}

In Visual Emotion Analysis (VEA), related work can be divided into several aspects according to different criteria.
Based on different psychological models, \ie Categorical Emotion States (CES) and Dimensional Emotion Space (DES), VEA can be grouped into classification task~\cite{chen2014deep,rao2016learning,yang2018visual,yang2018weakly,yang2018retrieving,yang2017joint,you2015robust, peng2016emotions, yang2021stimuli} and distribution learning task~\cite{yang2017learning, zhao2018discrete, zhao2018emotiongan, zhao2016continuous, zhao2017learning, yang2021circular}.
Oriented to distinct objects, VEA can be divided into personalized VEA~\cite{zhao2016predicting, zhao2018personality} and dominant VEA~\cite{chen2014deep,rao2016learning,yang2018visual,yang2018weakly,yang2018retrieving,yang2017joint,you2015robust,peng2016emotions}, which concentrate on one individual's emotion or the averaged one of the public.
Besides, there are also some surveys concerning this topic~\cite{joshi2011aesthetics, zhao2018affective, zhao2019affective, zhao2021affective}.
Our work focuses on dominant emotion classification problem.

\subsubsection{Traditional Methods}
\label{sec:traditional_methods}
Earlier works on VEA mainly focused on designing hand-crafted features to mine emotions from affective images.
Inspired by psychology and art theory, Machajdik~\etal~\cite{machajdik2010affective} extracted specific image features and combined them to predict emotions, which consisted of color, texture, composition and content.
Borth~\etal~\cite{borth2013large} introduced Adjective Noun Pairs (ANPs) and proposed a visual concept detector, \ie Sentibank, to filter out visual concepts strongly related to emotions from a semantic level.
Considering both bag-of-visual word representations and color distributions, Siersdorfer~\etal~\cite{siersdorfer2010analyzing} proposed an emotion analysis method based on information theory.
By leveraging object detection and concept modeling methods, Chen~\etal~\cite{chen2014object} first recognized the top six frequent objects,~\ie car, dog, dress, face, flower, food, and then modeled the concept similarity between those ANPs.
In order to understand the relationship between artistic principles and
emotions, Zhao~\etal~\cite{zhao2014exploring} proposed a method for both classification and regression tasks in VEA by extracting principle-of-art-based emotional features, including balance, emphasis, harmony, variety, gradation, and movement.
Besides, Zhao~\etal~\cite{zhao2014affective} extracted low-level generic and elements-of-art features, mid-level attributes and principles-of-art features, high-level semantic and facial features, followed by a multi-graph learning framework.
While these methods have been proven to be effective on several small-scale datasets, the hand-crafted features are still limited in covering all important factors in visual emotions.

\begin{figure*}
	\centering
	\includegraphics[width=\textwidth]{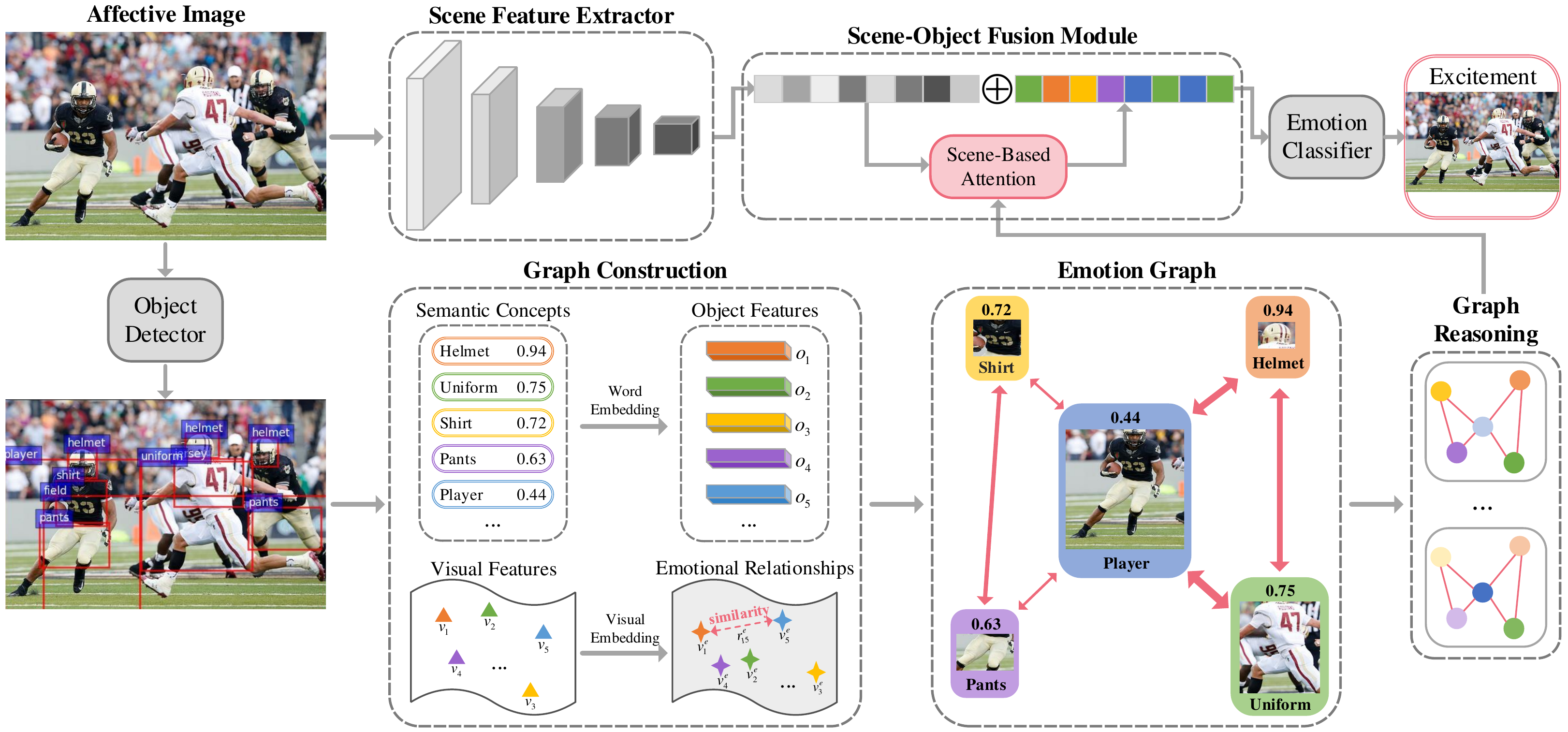}
	\vspace{-10pt}
	\caption{Framework of the proposed Scene-Object interreLated Visual Emotion Reasoning network (SOLVER).
		An object detector is first employed to extract semantic concepts and visual features of distinct objects (Sec.~\ref{sec:object_detection}).
		We then construct an Emotion Graph with objects as nodes and emotional relationships as edges (Sec.~\ref{sec:emotion_graph_construction}), and conduct GCN reasoning on it to yield emotion-enhanced object features (Sec.~\ref{sec:emotion_graph_reasoning}).
		Scene-Object Fusion Module is further designed by exploiting scene features to guide object fusion process with scene-based attention mechanism (Sec.~\ref{sec:scene_based_fusion_module}), where $\oplus$ denotes concatenation.
	}
	\vspace{-5pt}
	%	\Description{}
	\label{fig:network}
\end{figure*}

\subsubsection{Deep Learning Methods}
\label{sec:deep_learning_methods}
Recently, with the great success of deep learning networks, researchers in VEA have adopted Convolutional Neural Network (CNN) to predict emotions and have achieved significant progress.
Based on their previous SentiBank~\cite{borth2013large}, Chen~\etal~\cite{chen2014deep} implemented deep networks to construct a visual sentiment concept classification method named DeepSentiBank.
Leveraging half a million images labeled with website meta data, You~\etal~\cite{you2015robust} proposed a novel progressive CNN architecture (PCNN) to predict emotions.
Rao~\etal~\cite{rao2016learning} constructed a multi-level deep representation network (MldrNet), which extracted emotional features from  image semantics, aesthetics and low-level visual features simultaneously through multiple instance learning framework.
Earlier attempts simply implemented a general CNN to extract holistic features from affective images, neglecting the fact that visual emotions can also be evoked by local regions.
Different from previous methods, You~\etal~\cite{you2017visual} utilized attention mechanism to discover emotion-relevant regions, which serves as a prior attempt to focus on local regions in VEA.
Similarly, Yang~\etal~\cite{yang2018visual} constructed a local branch to discover affective regions by implementing the off-the-shelf detection tools. 
A weakly supervised coupled network (WSCNet)~\cite{yang2018weakly} was further proposed by Yang~\etal, which discovers emotion regions through attention mechanism and leverages both holistic and regional features to predict emotions in an end-to-end manner.
Besides, by employing deep metric learning, Yang~\etal~\cite{yang2018retrieving} proposed a multi-task deep framework for tackling both retrieval and classification tasks in VEA.
Zhang~\etal~\cite{zhang2019exploring} proposed a novel CNN model to extract and integrate content information as well as style information to infer visual emotions.
Considering different kinds of emotional stimuli, Yang~\etal~\cite{yang2021stimuli} proposed a stimuli-aware VEA network together with a hierarchical cross-entropy loss.
Most of the existing deep learning methods directly used holistic feature or regional features to predict visual emotions.
However, considering the complexity and abstractness involved in the cognitive process of human emotions, we argue that a direct mapping may underestimate the wide affective gap between low-level pixels and high-level emotions.

Psychological studies have demonstrated that visual emotions are evoked by the interactions between objects and objects as well as objects and scenes~\cite{brosch2010perception,frijda2009emotion,bar2004visual}.
Based on these observations, we propose a novel Scene-Object interreLated Visual Emotion Reasoning network (SOLVER) to predict visual emotions.
To mine the emotional relationships between different objects, we first construct an Emotion Graph and then conduct GCN reasoning on it to yield emotion-enhanced object features.
Besides, a Scene-Object Fusion Module is further proposed to effectively interrelate objects and scenes with a novel scene-based attention mechanism. 

\subsection{Object Detection}
\label{sec:object_detection_methods}
Object detection serves as a core problem in computer vision, which has shown dramatic progress recently.
Further, object detection methods have been implemented into various related tasks as a pre-processing step, including image captioning~\cite{yu2018topic,zhou2019re}, scene graph~\cite{xu2017scene,yang2018graph}, visual reasoning~\cite{xu2019reasoning,zhang2020temporal}, person re-identification~\cite{zhang2019multi, hao2019hsme},~\etc.
Object detection methods can be roughly divided into one-stage methods (\eg YOLO~\cite{redmon2016you}, SSD~\cite{liu2016ssd}) and two-stage methods (\eg R-CNN~\cite{girshick2014rich}, Fast R-CNN~\cite{girshick2015fast}, Faster R-CNN~\cite{ren2015faster}) according to whether region proposals are generated.
R-CNN was a pioneer work in two-stage detection methods, which boosted detection accuracy exceedingly by implementing deep networks for the first time.
Based on their previous work, Girshick proposed Fast R-CNN to reduce the redundant computation of feature extraction in R-CNN and further improve its accuracy and speed.
Region Proposal Network (RPN) was introduced by Faster R-CNN to replace the original time-consuming Selective Search, which not only unified object detection in an end-to-end network, but also improved its efficiency simultaneously.
Since Faster R-CNN is widely used for its great performance on both accuracy and speed~\cite{li2019transferable,hu2018relation,li2019visual,huang2020image,ji2020spatio}, we adopt Faster R-CNN to extract object features and then construct an Emotion Graph based on them.

\subsection{Graph-based Reasoning}
\label{sec:graph_based_reasoning_methods}
Recently, Graph Neural Networks (GNNs) have been proved to be an effective framework to exchange and propagate information through structured graphs, which have been widely used in various computer vision tasks~\cite{he2020mv,chen2019multi,chen2019graph}.
In Gated Graph Neural Network (GGNN)~\cite{li2015gated}, Li~\etal~adopted Gate Recurrent Units (GRUs) to update hidden states of graph nodes and implemented modern optimization techniques to yield output sequences.
Aiming to learn the graph-structured data via convolutional operations, Graph Convolutional Network (GCN)~\cite{kipf2016semi} was proposed as a scalable approach for semi-supervised classification.
In Graph Attention Networks (GAT)~\cite{velivckovic2017graph}, self-attention mechanism was introduced to update node features by attending over its neighbors, which was often used in bidirectional graph. 
In this paper, we apply GCN to conduct reasoning on the Emotion Graph, which correlates objects with their emotional relationships and yields emotion-enhanced object features.

\section{Methodology}
\label{sec:methodology}
In this section, we propose a novel Scene-Object interreLated Visual Emotion Reasoning network (SOLVER) to predict visual emotions through mining the interrelationships between objects and objects as well as objects and scenes.
Rather than adopting scene and object features directly, we develop a reasoning mechanism to infer deep emotional relationships between objects and scenes, which is helpful to bridge the existing affective gap.
Fig.~\ref{fig:network} shows the architecture of the proposed network.
We first employ an object detector, \ie Faster R-CNN, to extract semantic concepts and visual features of distinct objects (Sec.~\ref{sec:object_detection}).
After transforming and filtering those object features, we construct an Emotion Graph to depict emotional relationships between different objects (Sec.~\ref{sec:emotion_graph_construction}) and subsequently conduct GCN reasoning on it to correlate different objects with their pairwise relationships, yielding emotion-enhanced object features (Sec.~\ref{sec:emotion_graph_reasoning}).
Finally, we design a Scene-Object Fusion Module by exploiting scene features as guidance to fuse object features with scene-based attention mechanism, from which scene-object interrelationships are built simultaneously (Sec.~\ref{sec:scene_based_fusion_module}). 

\subsection{Object Detector}
\label{sec:object_detection}
We adopt Faster R-CNN~\cite{ren2015faster} as the object detector to select a set of candidate regions using ResNet-101~\cite{he2016deep} as backbone.
Our Faster R-CNN model is pre-trained on Visual Genome dataset~\cite{krishna2017visual}, which outputs attribute classes in addition to general object classes, providing more detailed information.
For example, object classes include: \textit{dog, building, mountains, woman},~\etc.,~while attribute classes include: \textit{white, large, rocky, beautiful},~\etc.
In the following process, we employ object classes alone, aiming to build up connections between emotions and objects. 
For simplicity, we only use the top-10 RoIs and further prove its effectiveness in Sec.~\ref{sec:hyper_parameter_analysis}.
After applying Faster R-CNN, each image can be represented as a set of object semantic concepts $S=\left\{ {{s}_{1}},{{s}_{2}},\dots,{{s}_{N}} \right\}$ with corresponding confidence scores $P=\left\{ {{p}_{1}},{{p}_{2}},\dots,{{p}_{N}} \right\}$, and a set of object visual features $\mathbf{V}=\left\{ {\mathbf{v}_{1}},{\mathbf{v}_{2}},\dots,{\mathbf{v}_{N}} \right\}$, in which ${\mathbf{v}_{i}}\in {{\mathbb{R}}^{d_1}}$, $N=10$, and $d_1=2048$.
Object semantic concepts $S$ are prediction results of the top-10 RoIs, which can be regarded as pseudo object-level labels since the involved visual emotion datasets only contain image-level labels.
Each confidence score ${p}_{i}$ indicates the degree of confidence towards its corresponding prediction result ${s}_{i}$.
For each selected region $i$, we extract the feature after the average pooling layer to serve as object visual feature ${v}_{i}$.

\subsection{Emotion Graph}
\label{sec:emotion_graph}
In this section, we build up an Emotion Graph and perform reasoning on it to mine the emotional relationships between different objects.
Based on the detection results of Faster R-CNN, we construct the Emotion Graph with objects as nodes and emotional relationships as edges, by transforming and filtering those detected object features in Sec.~\ref{sec:object_detection}.
Subsequently, we conduct GCN reasoning on the Emotion Graph, which propagates object information under different emotional relationships and eventually yields emotion-enhanced object features.
\subsubsection{Emotion Graph Construction}
\label{sec:emotion_graph_construction}
\begin{figure}
	\centering
	\includegraphics[width=0.46\textwidth]{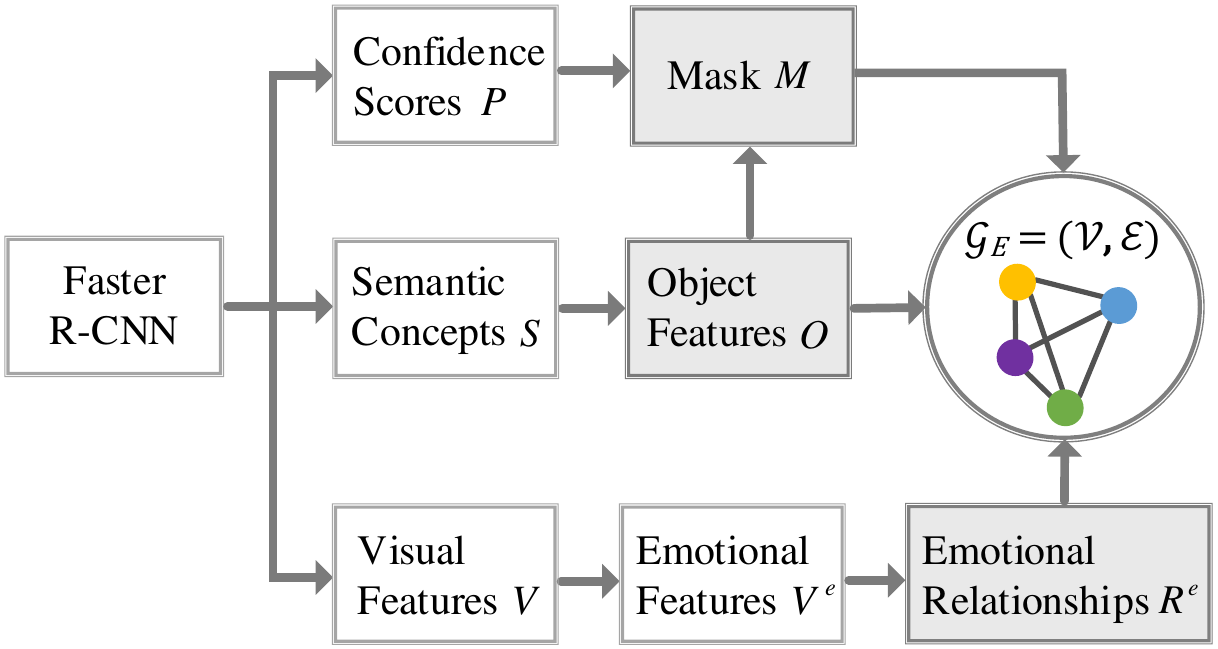}
	\caption{Construction of the Emotion Graph. Based on the detection results of Faster R-CNN, our Emotion Graph is constructed with objects as nodes and emotional relationships as edges. }
	\vspace{-10pt}
	%	\Description{}
	\label{fig:metho_1}
\end{figure}
We construct our Emotion Graph by setting objects as nodes and building up emotional relationships between them based on previous detection results as shown in Fig.~\ref{fig:metho_1}.
For word embedding, following the recent work~\cite{chen2019multi, huang2020image}, we employ Global Vectors for Word Representation (GloVe)~\cite{pennington2014glove}, an unsupervised learning algorithm to obtain vector representations for words.
Thus, semantic concepts $S=\left\{ {{s}_{1}},{{s}_{2}},\dots,{{s}_{N}} \right\}$ are embedded to semantic features $\mathbf O=\left\{ {\mathbf{o}_{1}},{\mathbf{o}_{2}},\dots,{\mathbf{o}_{N}} \right\}$, where ${\mathbf{o}_{i}}\in {{\mathbb{R}}^{d_2}}$ and $d_2=300$.
Notably, towards a specific object (\eg \textit{cat, tree, balloon}), visual features may vary from instance to instance while semantic features always remain unchanged.
Thus, instead of visual features $\mathbf V$, we adopt semantic features $\mathbf O$ as nodes to construct the Emotion Graph, aiming to map objects to emotions with one-to-one relationships.
Oppositely, we build up edges based on visual features $\mathbf V$, so as to depict diverse emotional relationships between distinct objects.
Considering the existing gap between semantics and emotions, we first embed visual features $\mathbf V$ from semantic space to emotional space using a linear function followed by a non-linear $\ell_2$-norm function:
\begin{align}
\label{eq:emo_emb}
\mathbf v_{i}^{e}=\ell_2({\mathbf {W}_{e}}  {\mathbf {v}_{i}}+{\mathbf{b}_{e}}),
\end{align}
which may bring a richer description towards emotional space.
In Eq.~\eqref{eq:emo_emb}, $i \in \{1,2,..., N\}$, ${\mathbf{W}_{e}}\in {{\mathbb{R}}^{N\times N}}$ is a learnable embedding matrix, and ${\mathbf{b}_{e}}\in {{\mathbb{R}}^{N}}$ is a learnable embedding bias.

Therefore, the emotional visual features are denoted as ${\mathbf{V}^{e}}=\left\{ {\mathbf{v}_{1}^{e}},{\mathbf{v}_{2}^{e}},\dots,{\mathbf{v}_{N}^{e}} \right\}$, in which ${\mathbf {v}_{i}^{e}}\in {{\mathbb{R}}^{d_1}}$.
Following~\cite{li2019visual,xu2019spatial}, we build up adjacency matrix by calculating affinity matrix to construct pairwise emotional relationships between objects for the Emotion Graph:
\begin{align}
\label{eq:vis_sim}
{{r}_{i,j}^{e}}=\phi {{\left(\mathbf v_{i}^{e} \right)}^{\mathrm{T}}}\!\cdot\! \varphi \left(\mathbf v_{j}^{e} \right),
\end{align}
where $i,j \in \{1,2,..., N\}$, $\mathbf v_{i}^{e}$, $\mathbf v_{j}^{e}$ represents two emotional visual features, and ${{r}_{i,j}^{e}}$ denotes the emotional relationships between them.
Notably, $\phi \left(\cdot \right) $ and $\varphi \left(\cdot \right)$ are two embedding functions with different parameters.
Since visual features ${\mathbf{V}^{e}}$ may vary greatly under the same emotion, we further embed them with different parameters to eliminate the bias of distinct features, hoping that they would be more comparable in the emotional space.
The settings in Eq.~\eqref{eq:vis_sim},  \ie two functions with different parameters, is further ablated in Section~\ref{sec:network_architecture_analysis}.

Our Emotion Graph is eventually constructed as ${\mathcal{G}_{E}}\!=\!\left( \mathcal V, \mathcal E \right)$
where nodes $\mathcal V$ denote the objects with their semantic features:
\begin{align}
\label{eq:node}
\mathbf O=\left\{ {\mathbf{o}_{1}},{\mathbf{o}_{2}},\dots,{\mathbf{o}_{N}} \right\},
\end{align}
and edges $\mathcal E$ represent the emotional relationships between different objects, which are described by affinity matrix:
\begin{align}
\label{eq:edge}
\mathbf {R}^{e}=\left[
\begin{matrix}
{r}_{1,1}^{e} & \cdots & {r}_{1,N}^{e}      \\
\vdots & \ddots & \vdots \\
{r}_{N,1}^{e} & \cdots & {r}_{N,N}^{e}      \\
\end{matrix}
\right],
\end{align}
which means there will be an edge with high affinity score connecting two objects if they have strong emotional relationships and are thus highly correlated.
In order to remove redundancy of object nodes, we set a confidence threshold of 0.3 for $P=\left\{ {{p}_{1}},{{p}_{2}},\dots,{{p}_{N}} \right\}$ to filter out less confident ones from the $N$ nodes, resulting in filtered nodes:
\begin{align}
\label{eq:O_m}
\mathbf O'= \left\{ {\mathbf{o}_{1}'},{\mathbf {o}_{2}'},\dots,{\mathbf {o}_{N}'}\right\},
\end{align}
\begin{align}
\label{eq:o_m}
{\mathbf {o}_{i}'}=\left\{\begin{matrix}
\!\!\mathbf {o}_{i}, \!\!\!\!&{{p}_{i}}\ge 0.3, &   \\
\!\!\ \mathbf{0},&{{p}_{i}}< 0.3, &
\end{matrix}\right.
\end{align}
where ${\mathbf{o}_{i}'}\in {{\mathbb{R}}^{d_2}}$ and $d_2=300$.
Notably, not only do we remove redundant nodes, but we also delete adjacent edges of these nodes through masking operations as shown in Eq.~\eqref{eq:m}, where $\mathrm {sgn(\cdot)}$ denotes sign function and $\mathrm {abs(\cdot)}$ denotes absolute value function.
After applying the above two functions, ${m}_{i,j}$ outputs either 1 or 0, suggesting whether or not edge ${r}_{i,j}^{e}$ still exists after the masking operation:
\begin{align}
\label{eq:M}
\mathbf M=\left[
\begin{matrix}
{m}_{1,1} & \cdots & {m}_{1,N}      \\
\vdots & \ddots & \vdots \\
{m}_{N,1} & \cdots & {m}_{N,N}      \\
\end{matrix}
\right],
\end{align}
\begin{align}
\label{eq:m}
{{m}_{i,j}}\!=\!\mathrm {sgn}\Big(\!\mathrm{abs}\big( \!\max \left( {\mathbf{o}_{i}'} \right) \!\times\! \max ( {\mathbf{o}_{j}'} ) \big)\! \Big),
\end{align}
where $i,j \in \{1,2,..., N\}$ and $\mathrm{max(\cdot)}$ takes the maximum element of ${\mathbf{o}_{i}'}$ and ${\mathbf{o}_{j}'}$.
We then apply mask matrix $\mathbf M$ to affinity matrix $\mathbf {R}^{e}$ and obtain the masked affinity matrix ${\mathbf{R}^{e}}'$ as
\begin{align}
\label{eq:edge_m}
{\mathbf{R}^{e}}'=\mathbf M\odot\mathbf {R}^{e}=\left[
\!\begin{matrix}
{{r}_{1,1}^{e}}' &\cdots & {{r}_{1,N}^{e}}'      \\
\vdots & \ddots & \vdots \\
{{r}_{N,1}^{e}}' & \cdots & {{r}_{N,N}^{e}}'      \\
\end{matrix}\!
\right],
\end{align}
where $\odot$ denotes the element-wise multiplication.
The validity of mask operation is further proved in Sec.~\ref{sec:network_architecture_analysis}.
In order to depict emotional relationships between objects, the Emotion Graph ${\mathcal{G}_{E}}\!=\!\left( \mathcal V, \mathcal E \right)$ is eventually built up with $\mathbf O'$ as nodes and ${\mathbf{R}^{e}}'$ as edges and further propagates information to yield emotion-enhanced features in Sec.~\ref{sec:emotion_graph_reasoning}.
\subsubsection{Emotion Graph Reasoning}
\label{sec:emotion_graph_reasoning}
In order to mine the emotional relationships between different objects, we apply Graph Convolutional Network (GCN)~\cite{kipf2016semi} to perform reasoning on the Emotion Graph, which is capable of exchanging and propagating information through structured graph.
In traditional GCN, as shown in Eq.~\eqref{eq:tra_GCN}, adjacency matrix is denoted as $\mathbf A$ while degree matrix is denoted as $\mathbf D$, where symmetric normalized Laplacian matrix ${{\widehat{\mathbf D}}^{-\frac{1}{2}}}\widehat{\mathbf A}{{\widehat{\mathbf D}}^{-\frac{1}{2}}}$ depicts the relationships between different nodes: 
\begin{align}
\label{eq:tra_GCN}
f\!\left( {{\mathbf H}^{\left( l \right)}},\mathbf A \right)=\sigma \left( {{\widehat{\mathbf D}}^{-\frac{1}{2}}}\widehat{\mathbf A}{{\widehat{\mathbf D}}^{-\frac{1}{2}}}{{\mathbf H}^{\left( l \right)}}{{\mathbf W}^{\left( l \right)}} \right),
\end{align}
where ${{\mathbf H}^{\left( l \right)}}$ denotes the node features of the \textit{l-th} layer and $\sigma(\cdot)$ denotes a nonlinear activation function.
Different from the supervised graph learning tasks, we need to establish a set of rules to calculate edges of the Emotion Graph, which is described in Eqs.~\eqref{eq:emo_emb}~\eqref{eq:vis_sim}~\eqref{eq:edge_m}.
Following similar update mechanism with Eq.~\eqref{eq:tra_GCN}, our GCN layer is defined as
\begin{align}
\label{eq:GCN}
f\left( {{\mathbf O'}^{\left( l \right)}},{{\mathbf R}^{e}}' \right)=\mathbf W_{r}^{\left( l \right)}\left( {\mathbf {R}^{e}}'{\mathbf {O}'^{\left( l \right)}}\mathbf W_{g}^{\left( l \right)} \right)+{\mathbf {O}^{\left( l \right)}},
\end{align}
where ${{\mathbf O'}^{\left( l \right)}}$ denotes the input node features of the \textit{l-th} layer and ${{\mathbf R}^{e}}'$ denotes the input edge features.
Notably, we add a residual block in our GCN following~\cite{li2019visual} to better maintain the original node features, drawing lessons from residual block in ResNet.
For the \textit{l-th} layer, $\mathbf W_{g}^{(l)}\in {{\mathbb{R}}^{N\times N}}$ is the weight matrix of GCN and $\mathbf W_{r}^{(l)}\in {{\mathbb{R}}^{N\times N}}$ is the weight matrix of the residual block.
It can be inferred from Eq.~\eqref{eq:GCN} that each node is updated based on both its neighbors and itself, from which object features propagate throughout the whole Emotion Graph under their emotional relationships.
We conduct reasoning on our Emotion Graph by applying several GCN layers to update object features iteratively:
\begin{align}
\label{eq:final_GCN}
{{\mathbf O'}^{\left( l \right)}}=f\left( {{\mathbf O'}^{\left( l-1 \right)}},{{\mathbf R}^{e}}' \right),
\end{align}
where $l \in \{1,2,..., L\}$.
The output of the final GCN layer ${{\mathbf O'}^{\left( L \right)}}\in {{\mathbb{R}}^{N\times d_2}}$ is regarded as emotion-enhanced object features, as object features are iteratively updated under their emotional relationships.
In our experiment, L is set to 4, which is further ablated in Sec.~\ref{sec:hyper_parameter_analysis}.
Therefore, we successfully construct the Emotion Graph and conduct reasoning on it to correlate objects with their emotional relationships.
\subsection{Scene-Object Fusion Module}
\label{sec:scene_based_fusion_module}
\begin{figure}
	\centering
	\includegraphics[width=0.4\textwidth]{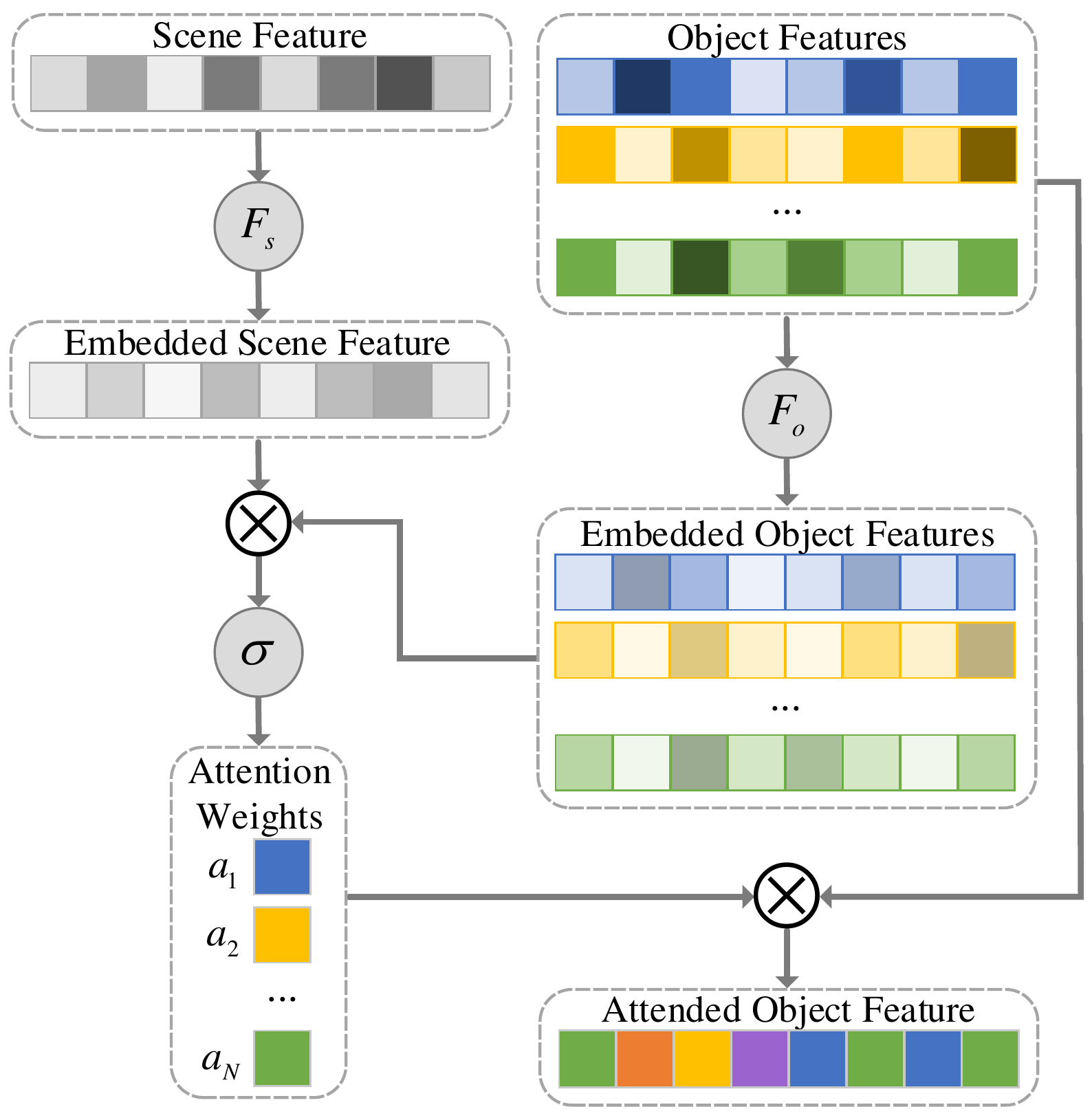}
	\caption{Scene-based attention mechanism. By exploiting scene features to guide object fusion process, scene-based attention mechanism interrelates scenes and objects with each other, where $\otimes$ represents matrix product.}
	%	\Description{}
	\label{fig:metho_2}
	\vspace{-10pt}
\end{figure}
In previous sections, we model the emotional relationships between distinct objects and yield emotion-enhanced object features.
Besides salient objects, scenes are regarded as another major stimulus in emotion evocation process, which largely affect the emotional tone of an image and thus cannot be ignored in VEA.
In this section, assuming that it is scenes that guide objects to evoke distinct emotions, we propose a novel scene-based attention mechanism to mine the scene-object interrelationships.
By implementing ResNet-50~\cite{he2016deep} as backbone, scene feature extractor takes an affective image as input and outputs with its corresponding scene feature, which is denoted as $\mathbf f_{sce} \in{{\mathbb{R}}^{d_1}} $ with $d_1=2048$.
In order to mine the deep interrelationships between scenes and objects, we propose a scene-based attention mechanism by exploiting scene features as guidance to fuse object features under different scene-object emotional relationships.
To be specific, we first project the scene feature $\mathbf f_{sce} \in{{\mathbb{R}}^{d_1}}$ and emotion-enhanced object features  $\mathbf O'={{\mathbf O'}^{\left( L \right)}} = \left\{{\mathbf {o}_{1}'},{\mathbf {o}_{2}'},\dots,{\mathbf {o}_{N}'}\right\},{\mathbf{o}_{i}'}\in {{\mathbb{R}}^{d_2}}$ into an embedding space to narrow the gap between scenes and objects, and calculate object attention weights by mining the emotional relationships between them:
\begin{align}
\label{eq:scene_att}
{{a}_{i}}=\sigma \big( {{F}_{s}}&\left( \mathbf f_{sce}  \right)\!\cdot\! {{F}_{o}}\left({\mathbf{o}_{i}}'\right)\!\big),
\end{align}
\begin{align}
\label{f_s_f_o}
{{F}_{s}}(\cdot)=\ell_2({\mathbf{W}_{s}} (\cdot)&),\  {{F}_{o}}(\cdot)=\ell_2({\mathbf{W}_{o}} (\cdot)),
\end{align}
where embedding functions are constructed by two learnable matrices $\mathbf{W}_{s}\in{{\mathbb{R}}^{d_1\times d_2}}$, $\mathbf{W}_{o}\in{{\mathbb{R}}^{d_2\times d_2}}$ followed by an $\ell_2$-norm function.
Besides, so as to normalize each attention weight ${{a}_{i}}$ to $[0,1]$, we apply Sigmoid function as the nonlinear activation function $\sigma(\cdot)$.
The closer relationship a specific object ${\mathbf{o}_{i}'}$ is to a scene $\mathbf f_{sce}$, the greater attention weight ${{a}_{i}}$ it will gain.
Considering that different interactions between objects and scenes may evoke different emotions, we employ their interrelationships, \ie the attention weights $A= \left\{{{a}_{1}},{{a}_{2}},\dots,{{a}_{N}}\right\}$, as guidance to fuse object features:
\begin{align}
\label{eq:att_sum}
\mathbf f_{obj}=\sum\limits_{i=1}^{N}{{{a}_{i}}{\mathbf {o}_{i}}'}.  
\end{align}

After applying attention weights to all object features, we obtain the attended object feature $\mathbf f_{obj}\in {{\mathbb{R}}^{d_2}}$ with $d_2=300$.
The overall process of the scene-based attention mechanism is shown in Fig.~\ref{fig:metho_2}.
Since both scenes and objects are indispensable in emotion evocation process, we further concatenate the scene feature $\mathbf f_{sce}\in {{\mathbb{R}}^{d_1}}$ with the object feature  $\mathbf f_{obj}\in {{\mathbb{R}}^{d_2}}$ as
\begin{align}
\label{eq:concate}
\mathbf f_{emo}=\mathrm {concate}\left[\mathbf f_{sce}, \mathbf f_{obj}\right].  
\end{align}

The concatenated emotion feature ${\mathbf f_{emo}}\in {{\mathbb{R}}^{d_1+d_2}}$ is then fed into the emotion classifier and a Softmax function successively:
\begin{align}
\label{eq:l_ce}
{{\mathcal{L}}_{CE}}=-\frac{1}{K} \sum\limits_{i=1}^{K} \sum\limits_{j=1}^{C}{{{y}}\left( i,j \right) \log \left( {{p}}\left( i,j \right) \right)},
\end{align}
\begin{align}
\label{eq:softmax}
{{p}}\left( i,j\left| \mathbf f_{emo},\mathbf W \right. \right)=\frac{\exp \left( {\mathbf{w}_{i,j}}\mathbf f_{emo} \right)}{\sum_{j=1}^{C}{\exp \left( {\mathbf{w}_{i,j}} \mathbf f_{emo} \right)}},
\end{align}
where $C$ denotes the number of emotion categories and $K$ denotes the number of affective images in a specific visual emotion dataset.
Moreover, $y$ represents emotion labels in the dataset while $p$ represents emotion prediction results of the proposed SOLVER.
$\mathbf W\in {{\mathbb{R}}^{(d_1+d_2) \times C}}$ is a learnable weight matrix in the emotion classifier, which is further optimized by the cross-entropy (CE) loss together with the whole network in an end-to-end manner.

\section{Experimental results}
\label{sec:experimental_results}
\subsection{Datasets}
\label{sec:datasets}

\begin{table}
	\centering
	%	\normalsize
	\caption{Statistics of the involved visual emotion datasets.}
	%	\vspace{-5pt}
	\label{tab:datasets}
	\renewcommand\arraystretch{1.18}
	\begin{tabular}{cp{1.4cm}<{\centering}p{1.4cm}<{\centering}p{1.4cm}<{\centering}p{1.4cm}}
		\toprule
		\toprule
		Dataset & {\# Images} & {\# Classes} & Type \\
		\midrule
		FI~\cite{you2016building} & 23,164 & 8 & Social \\
		Flickr~\cite{katsurai2016image} & 60,730 & 2 & Social\\
		Instagram~\cite{katsurai2016image} & 42,848 & 2 & Social\\
		EmotionROI~\cite{peng2016emotions} & 1,980 & 6 & Social\\
		Twitter I~\cite{you2015robust} & 1,269 & 2 & Social\\
		Twitter II~\cite{borth2013large} & 603 & 2 & Social\\
		ArtPhoto~\cite{machajdik2010affective} & 806 & 8 & Artistic\\
		IAPSa~\cite{lang1999international,mikels2005emotional} & 395 & 8 & Natural\\
		\bottomrule
		\bottomrule
	\end{tabular}
	\vspace{-10pt}
\end{table}

\begin{table*}
	\centering
	%	\normalsize
	\caption{Comparison with the state-of-the-art methods.\protect\\ Results are reported in classification accuracy (\%) on six visual emotion datasets.}
	\label{tab:SOTA}
	\renewcommand\arraystretch{1.18}
	\begin{tabular}{cp{1.6cm}<{\centering}p{1.6cm}<{\centering}p{1.6cm}<{\centering}p{1.6cm}<{\centering}p{1.6cm}<{\centering}p{1.6cm}<{\centering}p{1.6cm}}
		\toprule
		\toprule
		Method & FI & Flickr & Instagram & EmotionROI & Twitter I & Twitter II\\
		\midrule
		Sentibank~\cite{borth2013large} & 49.23 & 69.26 & 66.53 & 35.24 & 66.63 & 65.93 \\
		Zhao~\etal~\cite{zhao2014exploring} & 46.13 & 66.61 & 64.17 & 34.84 & 67.92 & 67.51\\
		DeepSentibank~\cite{chen2014deep} & 51.54 & 70.16 & 67.13 & 42.53 & 71.25 & 70.23\\
		Fine-tuned AlexNet~\cite{krizhevsky2012imagenet} & 59.85 & 79.73 & 77.29 & 44.19 & 75.20 & 75.63\\
		Fine-tuned VGG-16~\cite{simonyan2014very} & 65.52 & 80.75 & 78.72 & 49.75 & 78.35 & 77.31\\
		Fine-tuned ResNet-50~\cite{he2016deep} & 67.53  & 82.73 & 81.45 & 52.27 & 79.53 & 78.15\\
		MldrNet~\cite{rao2016learning} & 65.23 &--&--&--&--& --\\
		Sun~\etal~\cite{sun2016discovering} & -- & 79.85 & 78.67& --  & 80.33 & 78.97\\ 
		Yang~\etal~\cite{yang2017joint} & 67.48 &--&-- & 52.40 &--&--\\
		Yang~\etal~\cite{yang2018retrieving} & 67.64 &--&--&--&--& --\\
		WSCNet~\cite{yang2018weakly} & 70.07 & 81.36 & 81.81 & 58.25 & 84.25 & 81.35\\
		Zhang~\etal~\cite{zhang2019exploring} & 71.77 &--&--&--&--&--\\
		SOLVER (Ours) & \textbf{72.33} & \textbf{86.20} & \textbf{85.60}  & \textbf{62.12}& \textbf{85.43} & \textbf{83.19}\\
		\bottomrule
		\bottomrule
	\end{tabular}
	\vspace{-10pt}
\end{table*}

We evaluate the proposed SOLVER on eight public visual emotion datasets, including the Flickr and Instagram (FI)~\cite{you2016building}, Flickr, Instagram~\cite{katsurai2016image}, EmotionROI~\cite{peng2016emotions}, Twitter I~\cite{you2015robust}, Twitter II~\cite{borth2013large}, ArtPhoto~\cite{machajdik2010affective} and IAPSa~\cite{lang1999international,mikels2005emotional}.
The involved datasets can be roughly divided into large-scale datasets (\ie FI, Flickr, Instagram) and small-scale datasets (\ie EmotionROI, Twitter I, Twitter II, Artphoto, IAPSa), for which more details are shown in TABLE~\ref{tab:datasets}.

\textbf{FI.}
The FI dataset, with 23,164 images, is one of the largest well-labeled datasets, which is collected from the Flickr and Instagram by searching eight emotion categories as keywords, \ie\textit{Amusement, Anger, Awe, Contentment, Disgust, Excitement, Fear and Sad}. 
The collected images are then well-labeled by 225 Amazon Mechanical Turk (AMT) workers through keeping the weakly labels and their corresponding images with at least three of the five are agreed.  

\textbf{Flickr and Instagram.}
Using image ID or emotional words as query keywords, the Flickr and Instagram datasets are crawled from the internet, containing 60,730 and 42,848 affective images respectively.
Labeled by crowed-sourcing human annotation, these datasets provide sentiment labels with two emotion categories, \ie\textit{positive, negative}.
With approximately 100 thousand images in total, the Flickr and Instagram datasets provide us with a large dataset to train with deep learning methods.

\textbf{EmotionROI.}
The EmotionROI dataset contains 1,980 images with six emotion categories (\ie \textit{anger, disgust, fear, joy, sad, surprise}), which is a widely-used emotion prediction benchmark collected from Flickr. Besides, each image is also annotated with 15 bounding boxes as emotional regions, which act as pixel-level supervisions besides image-level labels.

\textbf{Twitter I.}
Collected from social websites, Twitter I contains 1,269 affective images in total.
Five AMT workers are recruited to generate sentiment labels for each candidate image in Twitter I. 

\textbf{Twitter II.}
Twitter II consists of 603 images downloaded from Twitter website, which is also labeled by AMT participants.
With sentiment labels, there are 470 positive images and 133 negative images in total.

\textbf{ArtPhoto.}
Rather than collecting images from social networks, ArtPhoto dataset are taken by professional artists, aiming to deliberately evoke certain emotions in their photos. 
Labeled with eight emotion categories, Artphoto dataset contains 806 images in total.

\textbf{IAPSa.} 
The International Affective Picture System (IAPS) dataset~\cite{lang1999international} is widely used in visual emotion analysis research. 
As a subset of the IAPS dataset, The IAPSa dataset~\cite{mikels2005emotional} contains 395 images and is labeled with eight emotion categories.

\subsection{Implementation Details}
\label{sec:implementation_details]}

Based on ResNet-50, our scene feature extractor is pre-trained on a large-scale visual recognition dataset, ImageNet~\cite{deng2009imagenet}.
We adopt Faster R-CNN as our object detector based on ResNet-101, which is pre-trained on Visual Genome dataset~\cite{krishna2017visual}.
The whole network, including the scene branch and the object branch, is then jointly trained in an end-to-end manner with affective datasets.
Following the same setting in~\cite{you2016building}, FI dataset is randomly split into training set (80\%), validation set (5\%) and testing set (15\%). 
Flickr and Instagram datasets are randomly split into training set (90\%) and testing set (10\%), which follows the same configuration in~\cite{katsurai2016image}.
Most of the small-scale datasets are split into training set (80\%) and testing set (20\%) randomly, except for those with specified training/testing separations~\cite{borth2013large}\cite{peng2016emotions}.
For training/validation/testing sets, we first resize each image to 480 on its shorter side and then crop it to 448$\times$448 randomly followed by a horizontal flip~\cite{he2016deep}. 
Our SOLVER is trained by the adaptive optimizer Adam~\cite{kingma2014adam}.
With a weight decay of 5e-5, the learning rate starts from 5e-5 and is decayed by 0.1 every 5 epochs, and the total epoch number is set to 50. 
Our framework is implemented using PyTorch~\cite{paszke2017automatic} and our experiments are performed on an NVIDIA GTX 1080Ti GPU.

\subsection{Comparison with the State-of-the-art Methods}
\label{sec:comparison_to_state_of_the_art}
\begin{figure*}
	\centering
	\includegraphics[width=0.95\linewidth]{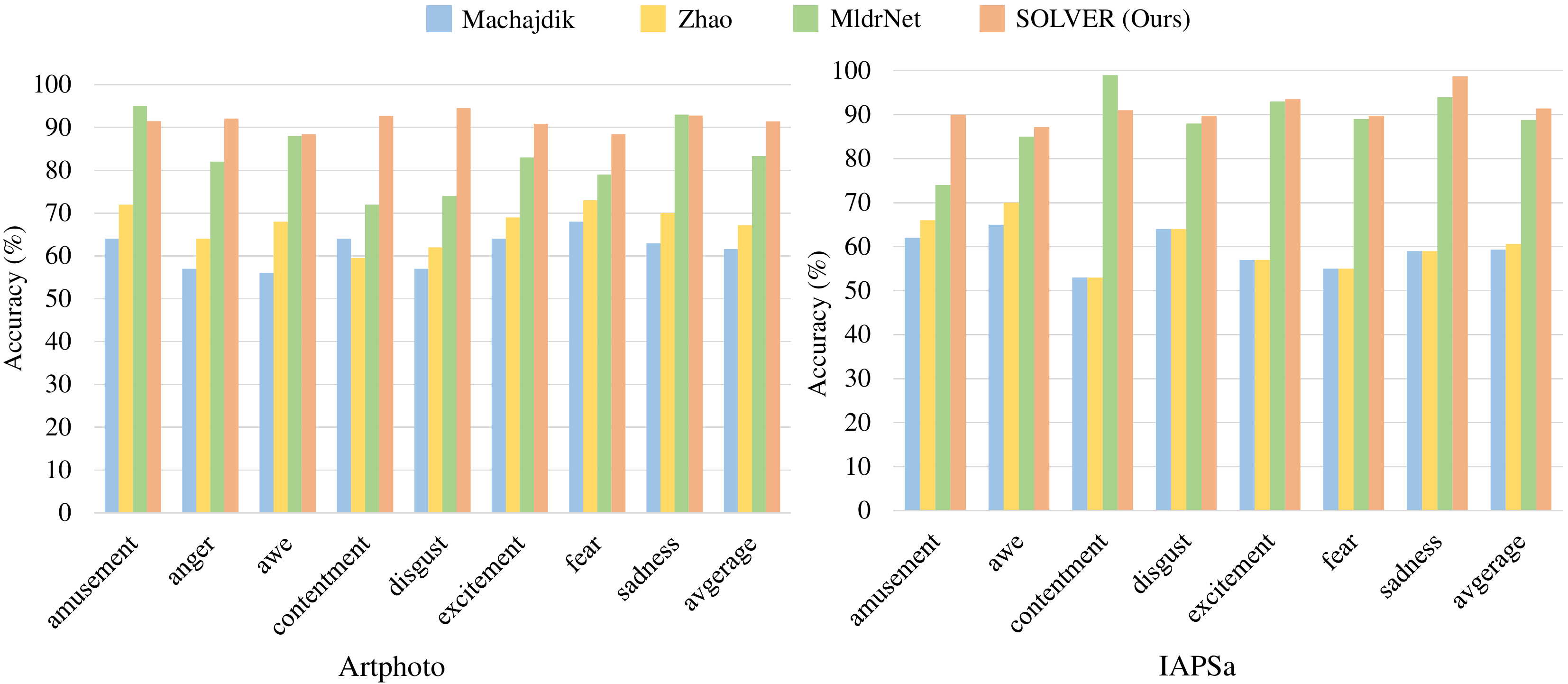}
	\vspace{-5pt}
	\caption{Classification results on Artphoto and IAPSa datasets, compared with the state-of-the-art methods.}
	%	\Description{}
	\label{fig:chart_1}
	\vspace{-10pt}
\end{figure*}

To evaluate the effectiveness of the proposed SOLVER, we conduct experiments compared with the state-of-the-art methods on eight visual emotion datasets, which are shown in TABLE~\ref{tab:SOTA} and Fig.~\ref{fig:chart_1}.

In TABLE~\ref{tab:SOTA}, we first compare our SOLVER with the state-of-the-art methods in classification accuracy on six visual emotion datasets, including FI, Flickr, Instagram, EmotionROI, Twitter I and Twitter II. which can be divided into traditional methods and deep learning ones.
For traditional methods, Sentibank~\cite{borth2013large} and Zhao~\etal~\cite{zhao2014exploring} adopted a set of emotion-related hand-crafted features, which were early attempts to explore the mysteries in VEA.
We also conduct experiments on several typical CNN backbones by fine-tuning the network parameters with visual emotion datasets, including AlexNet~\cite{krizhevsky2012imagenet}, VGG-16~\cite{simonyan2014very} and ResNet-50~\cite{he2016deep}.
Benefiting from its powerful representation ability, deep learning methods gained significant performance boosts compared with those traditional ones.
With coupled global-local branches, Yang \etal~\cite{yang2018weakly} proposed WSCNet and greatly improved the classification performance in VEA.
Zhang~\etal~\cite{zhang2019exploring} further boosted the classification performance by exploring content as well as style information to predict visual emotions.
Notably, the missing data in TABLE~\ref{tab:SOTA} is due to the lack of both classification results and open source codes.
We can infer from TABLE~\ref{tab:SOTA} that our SOLVER achieves greater performance boosts on large-scale datasets compared with those small-scale ones, as deep scene-object interrelationships can be better mined with larger datasets.
Overall, the proposed SOLVER outperforms the state-of-the-art methods by a large margin on six visual emotion datasets.
 
We further conduct detailed comparisons with the-state-of-the-art methods on Artphoto~\cite{machajdik2010affective} and IAPSa~\cite{lang1999international,mikels2005emotional}, as shown in Fig.~\ref{fig:chart_1}. 
Considering the limited and imbalanced data in the above datasets, we employ the ``one against all'' strategy to train our network following the previous method~\cite{machajdik2010affective} for fair comparison. 
Moreover, images in each category are randomly split into five batches and a 5-fold cross-validation is further implemented for emotion classification. 
We further remove the category of anger from IAPSa dataset following~\cite{machajdik2010affective,zhao2014exploring,rao2016learning}, as there are only eight samples in this category, where our SOLVER outperforms the state-of-the-art methods, including Machajdik\cite{machajdik2010affective}, Zhao~\etal\cite{zhao2014exploring} and MldrNet~\cite{rao2016learning}.
In addition to the optimal overall accuracy, our method is generally robust and applicable to each emotion with smaller accuracy deviation between categories.
The above analysis proves that our SOLVER is applicable on Artphoto and IAPSa datasets as well. 

The proposed SOLVER consistently outperforms the state-of-the-art methods on eight visual emotion datasets, which proves the effectiveness and robustness of our method.

\subsection{Ablation Study}
\label{sec:ablation_study}
Our SOLVER is further ablated to verify the validity of its network structure and the involved hyper-parameters.

\begin{table*}
	\centering
	%	\normalsize
	\caption{Ablation study of network structure on FI dataset.
	\label{tab:ablation}
	}	
	\renewcommand\arraystretch{1.18}
	\begin{tabular}{clc}
		\toprule
		\toprule
		Module & Ablated Combinations  & Acc (\%)\\
		\midrule
		\multirow{6}*{Emotion Graph}& single object & 34.98\\
		~ & multiple objects & 47.60\\
		~ & multiple objects + GCN + one embedding & 62.40\\
		~ & multiple objects + GCN + two embeddings & 64.47\\
		~ & multiple objects + GCN + mask + one embedding & 63.86\\
		~ & multiple objects + GCN + mask + two embeddings & 66.28\\
		\midrule
		\multirow{9}*{Scene-Object Fusion Module}& scene & 67.53\\
		~ & scene + single object & 68.13\\
		~ & scene + multiple objects & 70.20\\
%		~ & scene + multiple objects + GCN + mask + two embeddings & 71.41\\
		~ & scene + multiple objects + scene-based attention & 70.66\\
		~ & scene + multiple objects + GCN + one embedding + scene-based attention & 71.06\\
		~ & scene + multiple objects + GCN + two embedding + scene-based attention & 71.67\\
		~ & scene + multiple objects + GCN + mask + one embedding + scene-based attention & 71.93\\
		~ & scene + multiple objects + GCN + mask + two embeddings + scene-based attention & \textbf{72.33}\\
		\bottomrule
		\bottomrule
	\end{tabular}
	\vspace{-10pt}
\end{table*}

\subsubsection{Network Architecture Analysis}
\label{sec:network_architecture_analysis}

%\begin{table}
%	\centering
%%	\normalsize
%	\caption{Ablation study of network structure on FI dataset.}
%	\label{tab:ablation}
%%	\renewcommand\arraystretch{1.18}
%	\begin{tabular}{cccccc}
%		\toprule
%		\toprule
%		&&\multicolumn{2}{c}{Fusion Module}& \\
%		\cmidrule(lr){3-4}
%		Scene & Emotion Graph & GAP & Scene-Based Att & Acc (\%)\\
%		\midrule
%		&&$\checkmark$&& 47.60\\
%		$\checkmark$&&&& 67.53\\
%		&$\checkmark$&$\checkmark$ && 64.47\\
%		$\checkmark$&&$\checkmark$&& 70.20\\
%		$\checkmark$&&&$\checkmark$& 70.66\\
%		$\checkmark$&$\checkmark$&$\checkmark$&& 71.41\\
%		$\checkmark$&$\checkmark$&&$\checkmark$& \textbf{72.33}\\
%		\bottomrule
%		\bottomrule
%	\end{tabular}
%	\vspace{-10pt}
%\end{table}
As shown in TABLE~\ref{tab:ablation}, we conduct ablation study on FI dataset, aiming to verify the effectiveness of each proposed module.
Our SOLVER mainly consists of two modules: Emotion Graph (\ie object-object interaction) and Scene-Object Fusion Module (\ie scene-object interaction), as shown in the first column.
In each module, there are some detailed network designs, which are depicted as ablated combinations in the second column in TABLE~\ref{tab:ablation}.
In the Emotion Graph, we first conduct ablation studies concerning a single object and multiple objects, which indicates that multiple objects bring a performance boost compared with a single one.
After that, we introduce our GCN reasoning mechanism with detailed designs into comparisons, \ie one/two embedding(s), w/wo mask, suggesting that the object-object interactions indeed improve the performance in predicting emotions.
From the above experiments, it is obvious that two embedding functions with different parameters achieve better performance than one embedding function and mask operation can further bring a performance boost.
In the Scene-Object Fusion Module, it is obvious that both scene branch and scene-based attention mechanism make a great contribution to emotion classification, which suggests that it is the scene that guide object fusion process.  
From the above ablation studies, we can conclude that each detailed design of the proposed method is complementary and indispensable, which jointly contributes to the final result.
\begin{table}[h]
	%\arrayrulecolor{red}
	\centering
	%	\normalsize
	\caption{Ablation study of GCN layers (L) on FI dataset.}
	\label{tab:GCN_layers}
	
	%	\vspace{-5pt}
	\renewcommand\arraystretch{1.18}
	%	\resizebox{\textwidth}{5mm}{
	\setlength{\tabcolsep}{1.5mm}{
		\begin{tabular}{cccccccccc}
			\toprule
			\toprule
			L & 1 & 2 & 3 & 4 & 5 & 6 & 7 & 8 \\
			\midrule
			Acc (\%) & 71.03 & 71.49 & 72.04 & \textbf{72.33} & 72.36 & 72.01 & 72.13 & 71.58 \\
			\bottomrule
			\bottomrule
	\end{tabular}}
	\vspace{-10pt}
\end{table}

\begin{figure}
	\centering
	\includegraphics[width=0.95\linewidth]{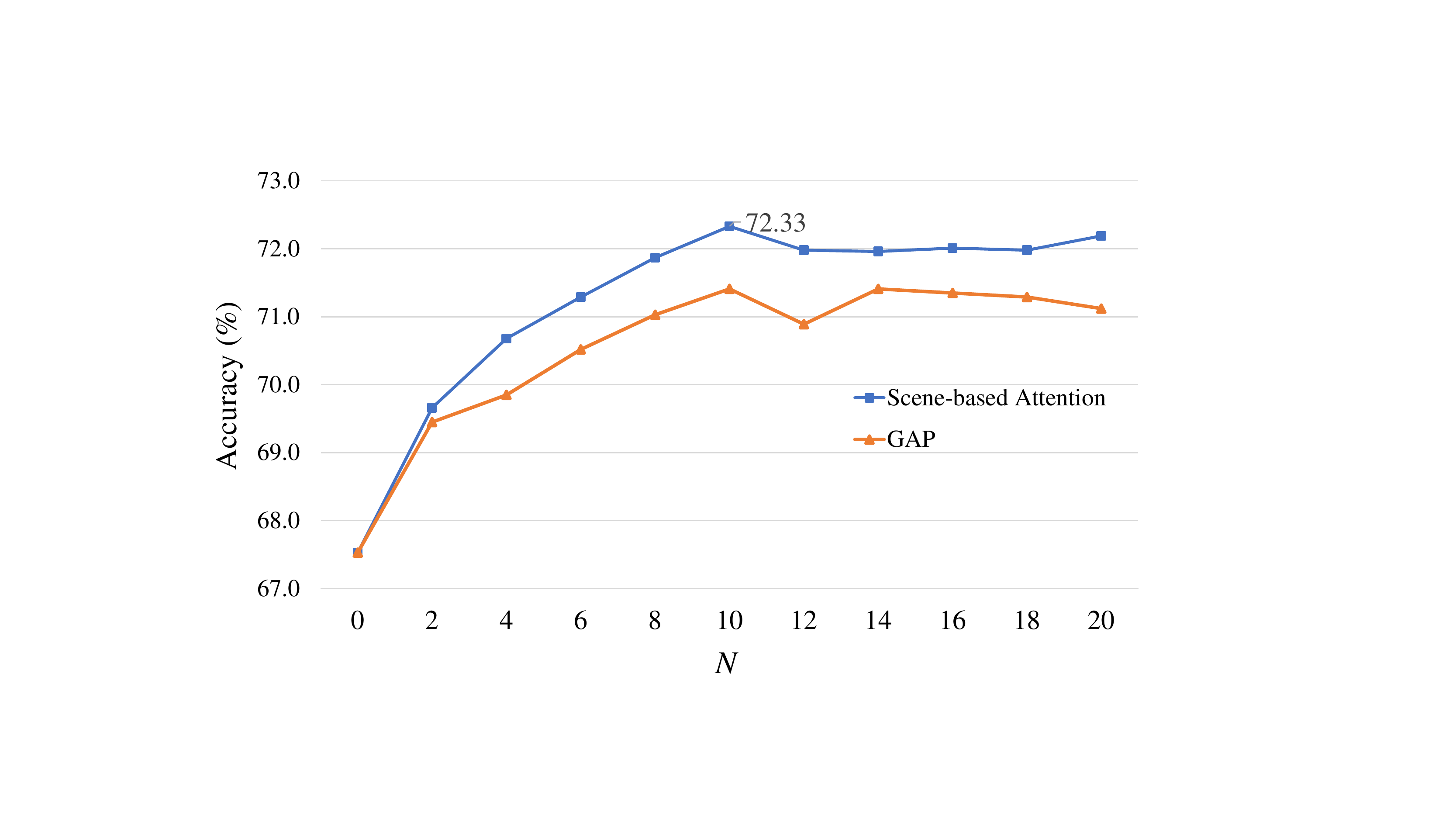}
	\vspace{-5pt}
	\caption{Hyper-parameter analysis of node number $N$ in the Emotion Graph.}
	%	\Description{}
	\label{fig:chart_2}
	\vspace{-10pt}
\end{figure}

\subsubsection{Hyper-Parameter Analysis}
\label{sec:hyper_parameter_analysis}
We conduct experiments to validate the choice of node number $N=10$ and the layer number $L=4$ in our Emotion Graph, as shown in Fig.~\ref{fig:chart_2} and TABLE~\ref{tab:GCN_layers}.
In general settings of Visual Genome dataset~\cite{krishna2017visual}, top-36 RoIs are selected from the pre-trained object detector.
However, we believe that such a large number of nodes can be redundant in our Emotion Graph and thus further conduct experiments to figure out the optimal $N$, balancing both accuracy and computational cost. 
To be specific, we design experiments on both scene-based attention mechanism and the GAP layer, aiming to reduce the randomness in a single experiment.
Setting $N=2$ as the step value, $N$ is varied from 0 to 20.
We find that the accuracy constantly grows as $N$ varies from 0 to 10, while it slightly drops after $N = 10$, due to the redundancy of RoIs in object detection.
It is worth mentioning that the growth of accuracy from 0 to 10 further proves that visual emotion analysis is not about a single object, but the interactions between multiple objects.
From the above analysis, we choose $N=10$ as the node number in our Emotion Graph.
For GCN layers, we ablate the number of GCN layers for better illustration.
Setting $L=1$ as the step value, L is varied from 1 to 8.
It is obvious that the accuracy consistently grows as the GCN layers increase, contributing to the excellent reasoning ability GCN owns.
However, the performance meets a bottleneck when the number of GCN layers is too large, which may be caused by over fitting of too many parameters.

\subsection{Visualization}
\label{sec:visualization}

The effectiveness of the proposed SOLVER has been quantitatively evaluated by comparing to the state-of-the-art methods and performing detailed ablation studies.
As we are motivated by the psychological evidences that emotions are evoked by the two categories of interactions, in this section, we try to figure out how objects and objects, objects and scenes are interrelated with each other by visualizing the intermediate process of the SOLVER.
To be specific, we visualize the emotional object concepts (Sec.~\ref{sec:emotional_objects}) and emotional object regions (Sec.~\ref{sec:emotional_regions}) for each category (\ie\textit{Amusement, Awe, Contentment, Excitement, Anger, Disgust, Fear and Sad}), which further validates the interpretability of our method and explores the mysteries of visual emotions.
\subsubsection{Emotional Object Concepts}
\label{sec:emotional_objects}
Considering that there exist relationships between objects, scenes and emotions, we visualize the top-10 emotional object concepts for each category.
After implementing Faster R-CNN on FI dataset, we first calculate the frequency ${{f}_{c,i}}$ for each object ${i \in \{1,2,..., I\}}$ in each emotion category ${c \in \{1,2,..., C\}}$, where $I$ denotes the overall object number and $C$ denotes the emotion categories in the whole dataset:
\begin{align}
\label{eq:frequency}
{{f}_{c,i}}&=\frac{{{N}_{c,i}}}{\sum_{i=1}^{I}{{{N}_{c,i}}}},
\end{align}
where ${{f}_{c,i}}\in \left[ 0,1 \right]$ and ${{N}_{c,i}}$ represents the count of object ${i}$ in category ${c}$.
Subsequently, we derive the scene-based attention coefficient ${{a}_{c,i,j}}$ from the trained model and calculate the averaged attention coefficient ${{a }_{c,i}}$ for each object ${i}$ in each emotion category ${c}$:
\begin{align}
\label{eq:att_average}
{{a }_{c,i}}=\frac{1}{{{N}_{c,i}}}\sum\limits_{j=1}^{{{N}_{c,i}}}{{{a }_{c,i,j}}},
\end{align}
where ${j \in \{1,2,..., {N}_{c,i}\}}$ denotes the \textit{j-th} instance of object $i$ in category $c$.
While object frequency ${{f}_{c,i}}$ represents the objects distribution of each emotion in the dataset, scene-based attention coefficient ${{a }_{c,i}}$ represents the correlations between objects and emotions learned by our SOLVER.
Taking both object frequency and scene-based attention coefficient into account, we obtain the weighted frequency ${{w}_{c,i}}$ for each object ${i}$ in each emotion category ${c}$:
\begin{align}
\label{eq:wei_frequency}
{{w}_{c,i}}={{f}_{c,i}}\times {{a }_{c,i}}.
\end{align}

\begin{figure*}
	\centering
	\includegraphics[width=\linewidth]{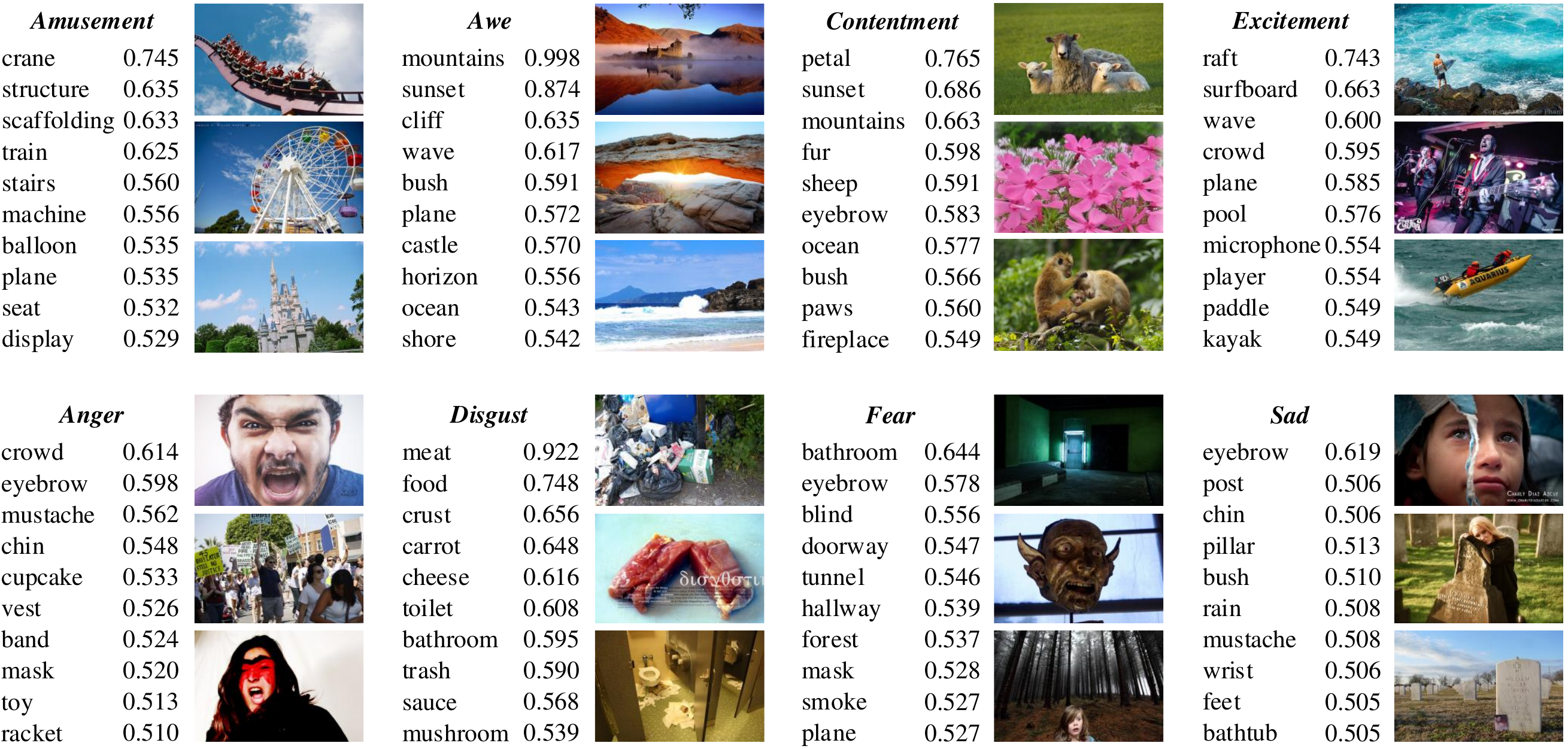}
	\caption{Visualization of emotional object concepts on FI. Each emotion is described with the top-10 emotional object concepts and their corresponding weighted frequencies. We further illustrate these emotional object concepts with three typical images from FI dataset.}
	%	\Description{}
	\label{fig:exp_1}
\end{figure*}

It is obvious that not all the objects are emotional, some non-emotional objects appear in every emotion category,~\eg \textit{man, woman, people,}~\etc. 
Thus, we adopt the TF-IDF~\cite{salton1973construction} technique to separate the emotion-specific objects from the non-emotional ones, where the importance of an object increases as it appears in a specific emotion category and decreases inversely with its appearance in the whole dataset.
As shown in Fig.~\ref{fig:exp_1}, we list the top-10 emotional object concepts for each emotion category with their corresponding weighted frequencies.
Besides, we present three typical images for each emotion to further illustrate these emotional object concepts in a concrete and vivid form.  
Taking \textit{awe} as an example, there are \textit{mountains, castle, ocean} in the first image, \textit{cliff, horizon, sunset} in the second one, and \textit{ocean, wave, shore} in the third one, which indicates that interactions between these emotional objects may evoke \textit{awe} to a large extent.
The most relevant concepts towards \textit{excitement} include \textit{raft, surfboard, plane, microphone, player,} \etc., which implies sports events or entertainments and consequently evokes \textit{excitement}.
\begin{figure*}
	\centering
	\includegraphics[width=\linewidth]{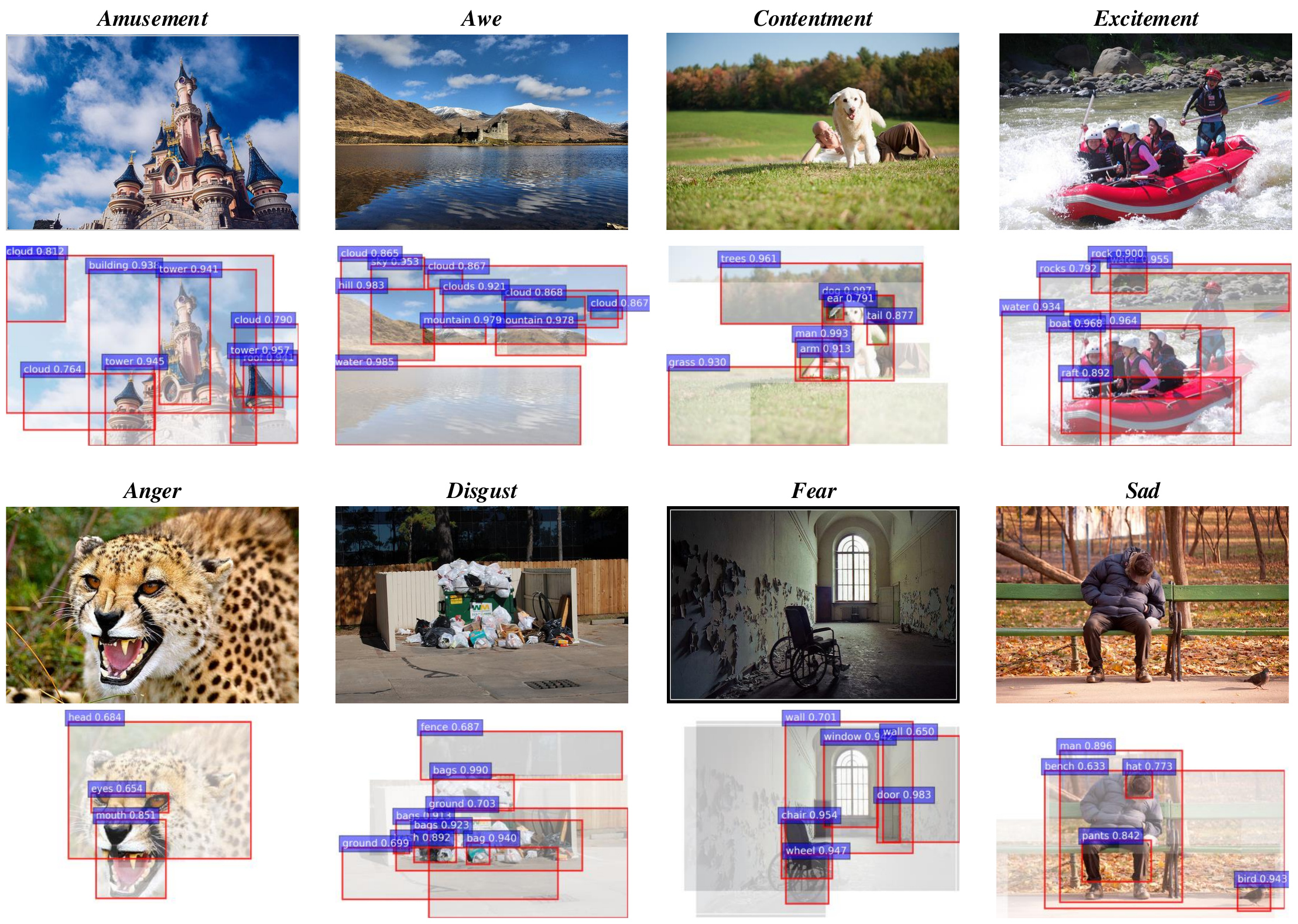}
	\caption{Visualization of emotional object regions on FI. Each emotion is represented with the input affective image and the output emotional object regions, together with their corresponding semantic concepts and attention weights.} 
	%	\Description{}
	\label{fig:exp_2}
\end{figure*}

\subsubsection{Emotional Object Regions}
\label{sec:emotional_regions}

Besides emotional object concepts, we also visualize emotional object regions by presenting one image for each emotion category as a representative.
As shown in Fig.~\ref{fig:exp_2}, after implemented the proposed SOLVER, an affective image (\ie the upper one) is turned into a set of emotional object regions (\ie the lower one) with corresponding semantic concepts and attention weights.
Notably, we first adopt object detectors to refine emotions to object level and then mine the interrelationships between objects and emotions with scene-based attention mechanism, which takes the scene feature as a guidance to fuse objects.
Attention weights reflect the relevance between objects, scenes and emotions.
The more an object is related to a specific emotion, the higher the attention weight will gain.
For example, in \textit{amusement}, \textit{tower (0.957), building (0.938)} get higher attention scores while \textit{cloud (0.764)} gets a lower score, as when \textit{tower} and \textit{building} comes together, they often indicate castle and amusement park, which surely bring people joy.
When people or other animals are \textit{angry}, the most striking part of their facial expression is an opened \textit{mouth}.
Thus, \textit{angry} is represented by an furious leopard with a distinguished opened \textit{mouth}, and accordingly our SOLVER focuses more on its \textit{mouth (0.851)} than its \textit{head (0.684), eyes (0.654)}. 

\subsection{Further Discussions}
\label{sec:fur_dis}
To validate the robustness of our method, we further extend experiments on three other potential datasets, \ie LUCFER~\cite{balouchian2019lucfer}, EMOTIC~\cite{kosti2017emotion}, and Flickr CC~\cite{borth2013large}, with more diverse emotional settings.
Meanwhile, we also notice some limitations of our method and discuss them for future work.
\subsubsection{Potential Datasets}
\label{po_datasets}
\textbf{LUCFER}~\cite{balouchian2019lucfer} contains over 3.6M images with 3-dimensional labels including emotion, context, and valence, which is currently the largest emotion recognition dataset.
With a sum of 18,316 images, \textbf{EMOTIC}~\cite{kosti2017emotion} serves as a pioneer dataset in emotion state recognition (ESR) field, which is labeled with both discrete categories (\ie 26 defined emotions) and continuous dimensions (\ie valence, arousal, and dominance (VAD)).
Containing 500,000 images in total, \textbf{Flickr CC}~\cite{borth2013large} is one of the largest visual sentiment datasets labeled with 1,553 fine-grained ANPs (Adjective Noun Pairs).
\subsubsection{Data Preprocessing}
\label{data-pre}
Since SOLVER is oriented to single-label emotion classification tasks, we preprocess the three potential datasets in distinct and reasonable ways to fit our task.
According to the rules in LUCFER, we degenerate the 275 fine-grained emotion-context pairs into eight basic emotions (\ie Anger, Anticipation, Disgust, Fear, Joy, Sadness, Surprise, and Trust) in Plutchik's wheel. 
Besides, we group the eight emotions into two sentiments (\ie positive, and negative) according to their polarities.
Following the same setting in~\cite{balouchian2019lucfer}, LUCFER is split into training set (80\%) and testing set (20\%).
Each image in EMOTIC is annotated with multiple emotion labels, which is different from our single-label classification task.
Besides, it is irrational to degrade a multi-label task to a single-label one.
Fortunately, valence in VAD measures the positive degree of an emotion, where $V\in [5,10]$ corresponds to positive and $V\in [0,5]$ negative.
Thus, we degenerate valence to sentiment labels (\ie positive, and negative) according to psychological model.
EMOTIC is split into training set (70\%), validation set (10\%), and testing set (20\%)~\cite{kosti2017emotion}.
Based on Visual Sentiment Ontology (VSO)~\cite{borth2013large}, we degenerate its 1,553 ANPs into sentiment labels (\ie positive, and negative).
The Flickr CC dataset is split into training set (80\%) and testing set (20\%), which follows the same setting in~\cite{borth2013large}.

\begin{table}
	\centering
	%	\normalsize
		\caption{Classification accuracy (\%) on three potential datasets \protect\\ using models trained on FI.}
		%	\vspace{-5pt}
		\label{tab:add_FI}
	\renewcommand\arraystretch{1.18}
	\begin{tabular}{cp{1.4cm}<{\centering}p{1.4cm}<{\centering}p{1.4cm}<{\centering}p{1.4cm}}
		\toprule
		\toprule
		Dataset & Scene & Object & SOLVER \\
		\midrule
		LUCFER-8~\cite{balouchian2019lucfer} & 12.98 & 17.06 & 14.69 \\
		LUCFER-2~\cite{balouchian2019lucfer} & 65.46 & 67.19 & 69.38\\
		EMOTIC-2~\cite{kosti2017emotion} & 36.11 & 35.99 & 36.13\\
		Flickr CC-2~\cite{borth2013large} & 37.05 & 36.45 & 36.86\\
		\bottomrule
		\bottomrule
	\end{tabular}
	\vspace{-5pt}
\end{table}

\begin{table}
	\centering
	%	\normalsize
	\caption{Classification accuracy (\%) on three potential datasets \protect\\ using models trained on themselves.}
	%	\vspace{-5pt}
	\label{tab:add_ALL}
	\renewcommand\arraystretch{1.18}
	\begin{tabular}{cp{1.4cm}<{\centering}p{1.4cm}<{\centering}p{1.4cm}<{\centering}p{1.4cm}}
		\toprule
		\toprule
		Dataset & Scene & Object & SOLVER \\
		\midrule
		LUCFER-8~\cite{balouchian2019lucfer} & 72.04 & 71.25 & 75.59\\
		LUCFER-2~\cite{balouchian2019lucfer} & 87.94 & 86.84 & 90.55\\
		EMOTIC-2~\cite{kosti2017emotion} & 63.47 & 62.22 & 65.07\\
		Flickr CC-2~\cite{borth2013large} & 69.74 & 68.27 & 71.68\\
		\bottomrule
		\bottomrule
	\end{tabular}
	\vspace{-5pt}
\end{table}

\subsubsection{Classification Accuracy}
\label{add_cla_acc}
We conduct experiments in classification accuracy on three potential datasets in TABLE~\ref{tab:add_FI} and TABLE~\ref{tab:add_ALL}.
Specifically, results using models trained on FI dataset are reported in TABLE~\ref{tab:add_FI}, and results using models trained on different datasets themselves are shown in TABLE~\ref{tab:add_ALL}.
In particular, LUCFER is conducted with two experimental settings, eight emotions (\ie LUCFER-8) and two sentiments (\ie LUCFER-2), for richer comparisons and analyses.
In order to validate the effectiveness of our method and to explore the emotional diversity in different datasets, we ablated our experiments to three settings including Scene branch, Object branch, and SOLVER.

\begin{figure*}
	\centering
	\includegraphics[width=\linewidth]{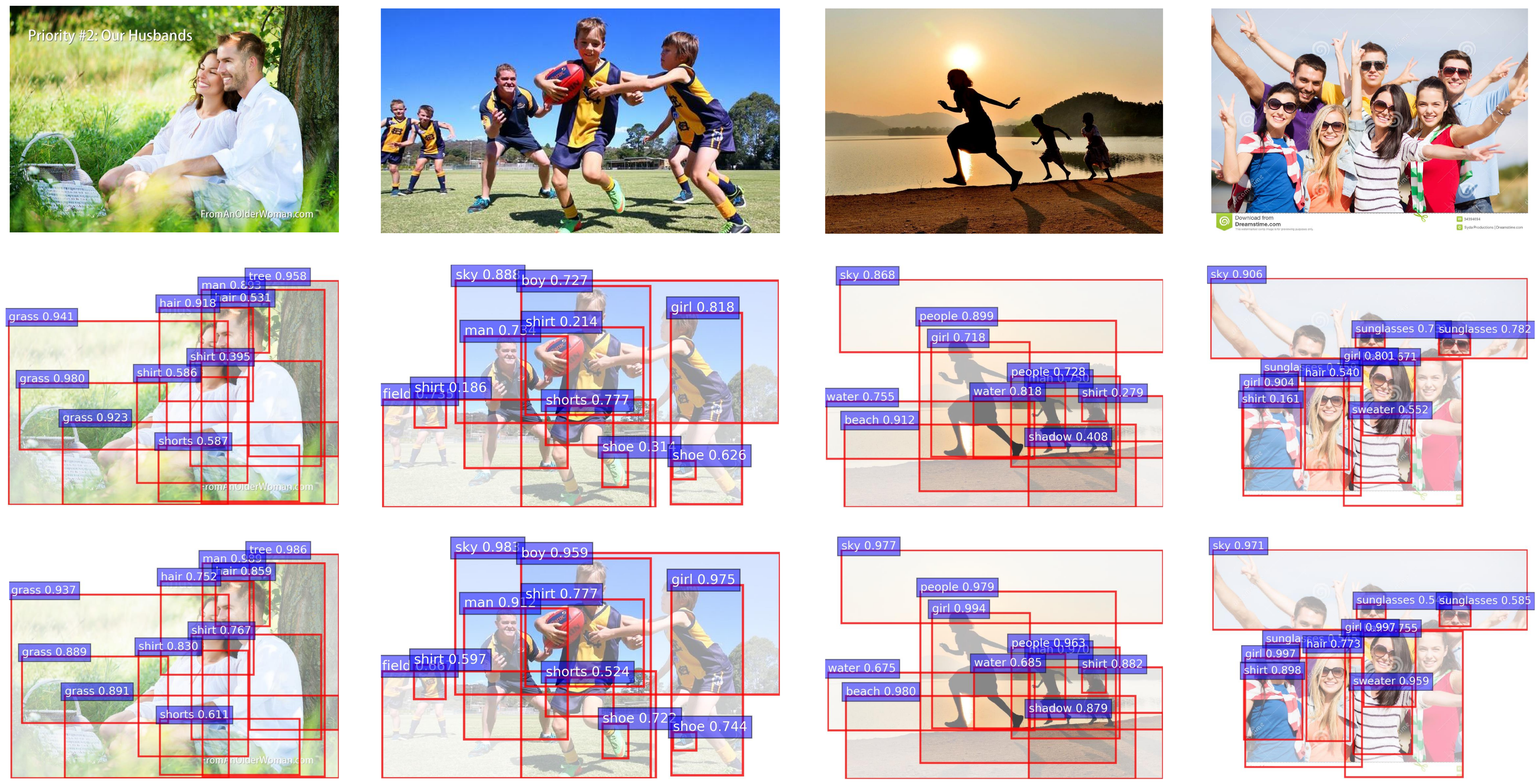}
	\caption{Visualization of emotional object regions on LUCFER. Each image (\ie the first row) is presented with two sets of emotional object regions (\ie the second and third rows). Specifically, the second row uses model trained on FI while the model in the third row is trained on LUCFER itself.}
	%	\Description{}
	\label{fig:add_lucfer_visual}
\end{figure*}

In TABLE~\ref{tab:add_FI}, it can be noticed that the FI trained model achieves relatively good results on LUCFER-2 (69.38\%), while meets unsatisfactory results on LUCFER-8, EMOTIC-2 and Flickr CC-2 (14.69\%, 36.13\%, and 36.86\%).
Dataset shift is a common problem in machine learning. 
Since emotions are complex and ambiguous to be described and labeled, this gap becomes even larger. 
Besides, VEA aims at finding out how people feel emotionally towards different visual stimuli, while ESR focuses on recognizing peoples' emotional states from their frames.
The mismatches between these two areas may lead to a performance degradation in TABLE~\ref{tab:add_FI}. 
%In general, 2-sentiment classification tasks perform better than 8-emotion ones, owing to the lower classification difficulty with more distinguishable features and fewer categories. 
Compared with the other 2-sentiment tasks, LUCFER-2 achieves quite a good result, which attributes to the well-constructed dataset with more emotional stimuli and less noise. 
It is obvious that Object branch performs generally better than Scene branch on LUCFER. 
The possible reason may be that there are distinct and obvious objects conveying emotions, while scene features are not clearly distinguishable to separate different emotions apart.
The three results on EMOTIC dataset are close to each other, which indicates that our SOLVER failed to find key features related to ESR.
Moreover, we guess that the key features concerning ESR may be facial expressions or body language, which our method does not involve.
Similarly, SOLVER performs poorly on Flickr CC, resulting from its very large data scale compared with FI.
Besides, most images in Flickr CC contain a distinct object or a simple scene, which our SOLVER may fail to match with.

For fair comparisons with datasets in Sec~\ref{sec:comparison_to_state_of_the_art}, we also report the results using models trained on three potential datasets and tested on themselves in TABLE~\ref{tab:add_ALL}.
Compared with TABLE~\ref{tab:add_FI}, it is obvious that the corresponding results in TABLE~\ref{tab:add_ALL} have significant performance improvements, which are comparable to the results in TABLE~\ref{tab:SOTA}.
In TABLE~\ref{tab:add_ALL}, it is undoubted that LUCFER is a well-labeled dataset, whose 8-emotion accuracy (75.59\%) is higher than others' 2-sentiment ones (65.07\%, and 71.68\%).
TABLE~\ref{tab:add_ALL} shows that SOLVER constantly outperforms the other ablated branches on all experimental settings, which further validates the effectiveness and the robustness of each proposed branch. 
%Comparing the ablated branches, \ie Scene branch and Object branch, it is obvious that Scene branch outperforms Object branch, which may contribute to the rich emotional information within the scene.

From TABLE~\ref{tab:add_FI} and TABLE~\ref{tab:add_ALL}, we can conclude that SOLVER achieves competitive results on three potential datasets.
Meanwhile, we also noticed some limitations of our method. 
Firstly, when facing ESR tasks, SOLVER fails to leverage the facial expressions as well as body language of people within the images, resulting in degraded performance. 
Secondly, SOLVER is proposed to mine the emotional correlations between objects and scenes, which fails to deal with the situation of a single object or a simple scene.

\begin{figure*}
	\centering
	\includegraphics[width=\linewidth]{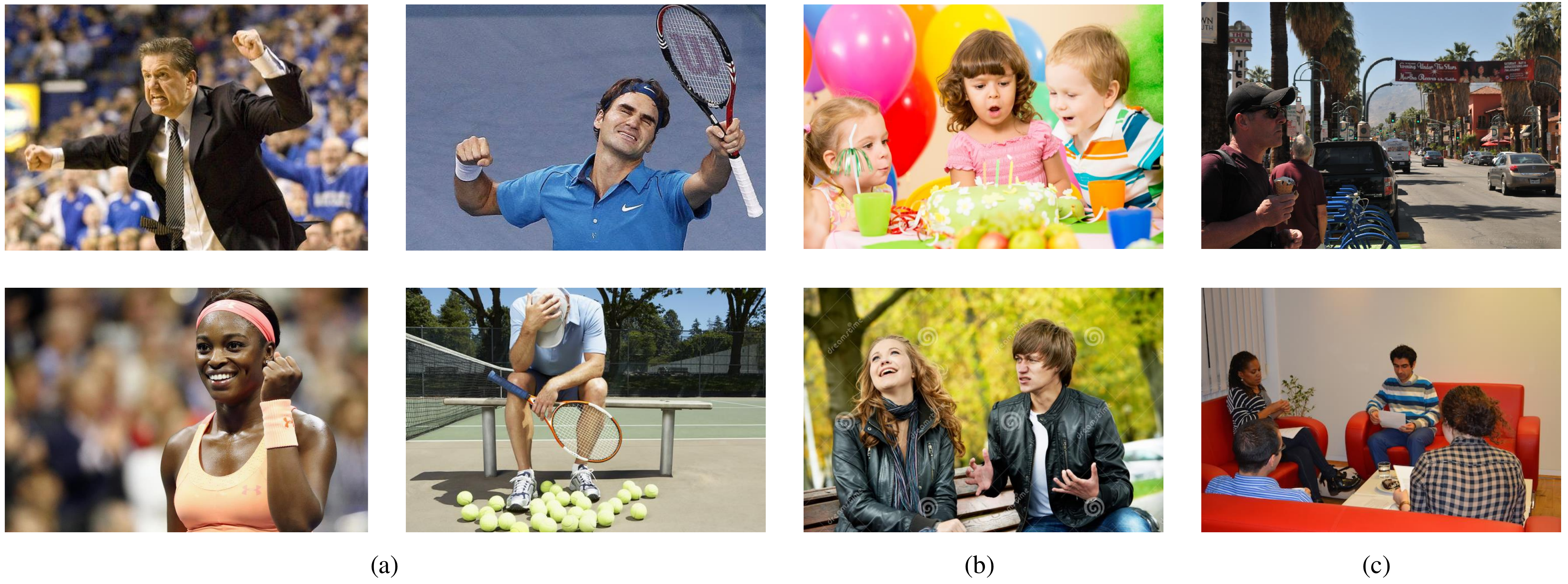}
	\caption{Failure cases on LUCFER. (a) Rather than specific scenes (\ie \textit{stadium} in the first column) or objects (\ie \textit{tennis} in the second column), emotions are largely evoked by facial expressions and body language of people within the image. (b) Different people in an image do not necessarily share one emotion. (c) Some images do not necessarily evoke strong emotions.}
	%	\Description{}
	\label{fig:add_lucfer_failure}
\end{figure*}

\subsubsection{Visualizations}
\label{add_visual}
In addition to classification accuracy, we also visualize emotional object regions on potential datasets from different trained models (\ie FI and themselves).
In Fig.~\ref{fig:add_lucfer_visual}, we present the visualizations of emotional object regions on LUCFER. 
While the first row represents the input images, the second row uses model trained on FI and the third row is trained on LUCFER itself.
By analyzing the difference in attention weights, we discover some possible reasons for the performance improvements in TABLE~\ref{tab:add_ALL} compared with TABLE~\ref{tab:add_FI}. 
Take the first image as an example, while the second row focuses more on \textit{grass, tree}, and \textit{other objects}, the third row attends more weights on \textit{man, shirt,} and \textit{hair}. 
It is obvious that FI trained model concentrate more on objects unrelated to people while LUCFER trained model emphasizes more on human-centered ones, which further shows the robustness of our method. 
As ESR datasets are constructed by images all containing people, it is understandable that increased attention weights on people and their related objects may bring a performance boost in such datasets.
As is mentioned above, there are some similarities in VEA and ESR tasks yet much more differences in details.
%As is mentioned above, there are some similarities in VEA and ESR tasks yet much more differences in details. where VEA aims at finding the viewer's emotion while ESR focuses on the people's emotion within the image.
%In VEA, we can successfully infer visual emotions from the interactions between objects and scenes by implementing SOLVER. 
%However, in ESR, the specific emotion conveyed by an image is largely determined by the facial expressions and body language of people within it.
The visualizations also prove that emotions are always evoked by people within the images in ESR datasets, especially their facial expressions and body language, which is different from our VEA task.

\subsubsection{Failure Cases}
\label{add_failure}
%Although SOLVER achieves a comparable result on potential datasets, there are still some failure cases. 
%On the one hand, ESR datasets are mismatched with VEA problem. 
%On the other hand, emotions are complex and abstract, which may be difficult to fully considered by a single method.
%In LUCFER, failure cases are summarized into three types in Fig.~\ref{fig:add_lucfer_failure}.
Fig.~\ref{fig:add_lucfer_failure} shows three types of failure cases on LUCFER.
In Fig.~\ref{fig:add_lucfer_failure} (a), two images in the first column share the same scene (\ie \textit{stadium}), and two images in the second column share the same object (\ie \textit{tennis}). 
However, under the same scene/object, emotions are varying according to different facial expressions and body language. 
Take first row as an example, by observing the \textit{waving fists} and the \textit{confident smile}, the coach is \textit{angry} with his players while the player is \textit{satisfied} with herself
Since SOLVER mines emotions from the interactions between objects and scenes, facial expressions and body language are not considered in our method. 
This failure is mainly caused by the mismatch between VEA and ESR, which will be considered in our future work. 
In Fig.~\ref{fig:add_lucfer_failure} (b), we can see that people in an image do not necessarily share the same emotion. 
Take the second row as an example, the woman seems \textit{happy} while the man is \textit{angry}. 
%As our task aims at predicting the specific emotion evoked by an image, it is hard to decide which one is the dominant emotion in such cases. 
While ESR concentrates on the emotions of characters within the image, our VEA focuses on the emotions of viewers outside the image. 
This failure also results from the mismatch between two tasks. 
Besides, there are also some images where emotions may be weak and obscure as in Fig.~\ref{fig:add_lucfer_failure} (c). 
In other words, these images hardly arouse any of our emotions. 
This failure is universal in both tasks. 
We may find a reasonable way to separate emotional images and emotionless images in our future work.

\section{Conclusion}
\label{sec:conclusion}
We have proposed a Scene-Object interreLated Visual Reasoning network (SOLVER) to mine emotions from the interactions between objects and objects as well as objects and scenes.
We first constructed an Emotion Graph based on detected features and conducted GCN reasoning on it, aiming to extract the emotional relationships between different objects.
Besides, we proposed a Scene-Object Fusion Module to fuse objects with the guidance of scene-based attention mechanism.
Extensive experiments and comparisons have shown that the proposed SOLVER consistently outperforms the state-of-the-art methods on eight public visual emotion datasets.
Notably, visualization results on emotional object concepts and regions not only proved the interpretability of our network, but also offered new insight to explore the mysteries in visual emotion analysis.
We further extended our experiments on three other potential datasets, where we validated the effectiveness and robustness of our method with more diverse emotional settings.
Additionally, we also noticed some limitations of our method.
Since SOLVER aims at mining emotions from the interactions between objects and scenes, it may fail to deal with situations including images with a single object/scene and drawings without objects/scenes.
In ESR datasets, though scenes and objects are still important in predicting emotions, human-centered attributes, \ie facial expressions and body language, play a leading role that cannot be ignored.
These limitations will be considered in a unified network for a better performance in our future work.

% trigger a \newpage just before the given reference
% number - used to balance the columns on the last page
% adjust value as needed - may need to be readjusted if
% the document is modified later
%\IEEEtriggeratref{8}
% The "triggered" command can be changed if desired:
%\IEEEtriggercmd{\enlargethispage{-5in}}

% references section

% can use a bibliography generated by BibTeX as a .bbl file
% BibTeX documentation can be easily obtained at:
% http://mirror.ctan.org/biblio/bibtex/contrib/doc/
% The IEEEtran BibTeX style support page is at:
% http://www.michaelshell.org/tex/ieeetran/bibtex/
\bibliographystyle{IEEEtran}
\bibliography{mybibfile_revised}
% argument is your BibTeX string definitions and bibliography database(s)
%\bibliography{IEEEabrv,../bib/paper}
%
% <OR> manually copy in the resultant .bbl file
% set second argument of \begin to the number of references
% (used to reserve space for the reference number labels box)

% biography section
% 
% If you have an EPS/PDF photo (graphicx package needed) extra braces are
% needed around the contents of the optional argument to biography to prevent
% the LaTeX parser from getting confused when it sees the complicated
% \includegraphics command within an optional argument. (You could create
% your own custom macro containing the \includegraphics command to make things
% simpler here.)
%\begin{IEEEbiography}[{\includegraphics[width=1in,height=1.25in,clip,keepaspectratio]{mshell}}]{Michael Shell}
% or if you just want to reserve a space for a photo:

%\begin{IEEEbiography}{Michael Shell}
%Biography text here.
%\end{IEEEbiography}
%
%% if you will not have a photo at all:
%\begin{IEEEbiographynophoto}{John Doe}
%Biography text here.
%\end{IEEEbiographynophoto}
%
%% insert where needed to balance the two columns on the last page with
%% biographies
%%\newpage
%
%\begin{IEEEbiographynophoto}{Jane Doe}
%Biography text here.
%\end{IEEEbiographynophoto}

% You can push biographies down or up by placing
% a \vfill before or after them. The appropriate
% use of \vfill depends on what kind of text is
% on the last page and whether or not the columns
% are being equalized.

%\vfill

% Can be used to pull up biographies so that the bottom of the last one
% is flush with the other column.
%\enlargethispage{-5in}

\begin{IEEEbiography}[{\includegraphics[width=1in,height=1.25in,clip,keepaspectratio]{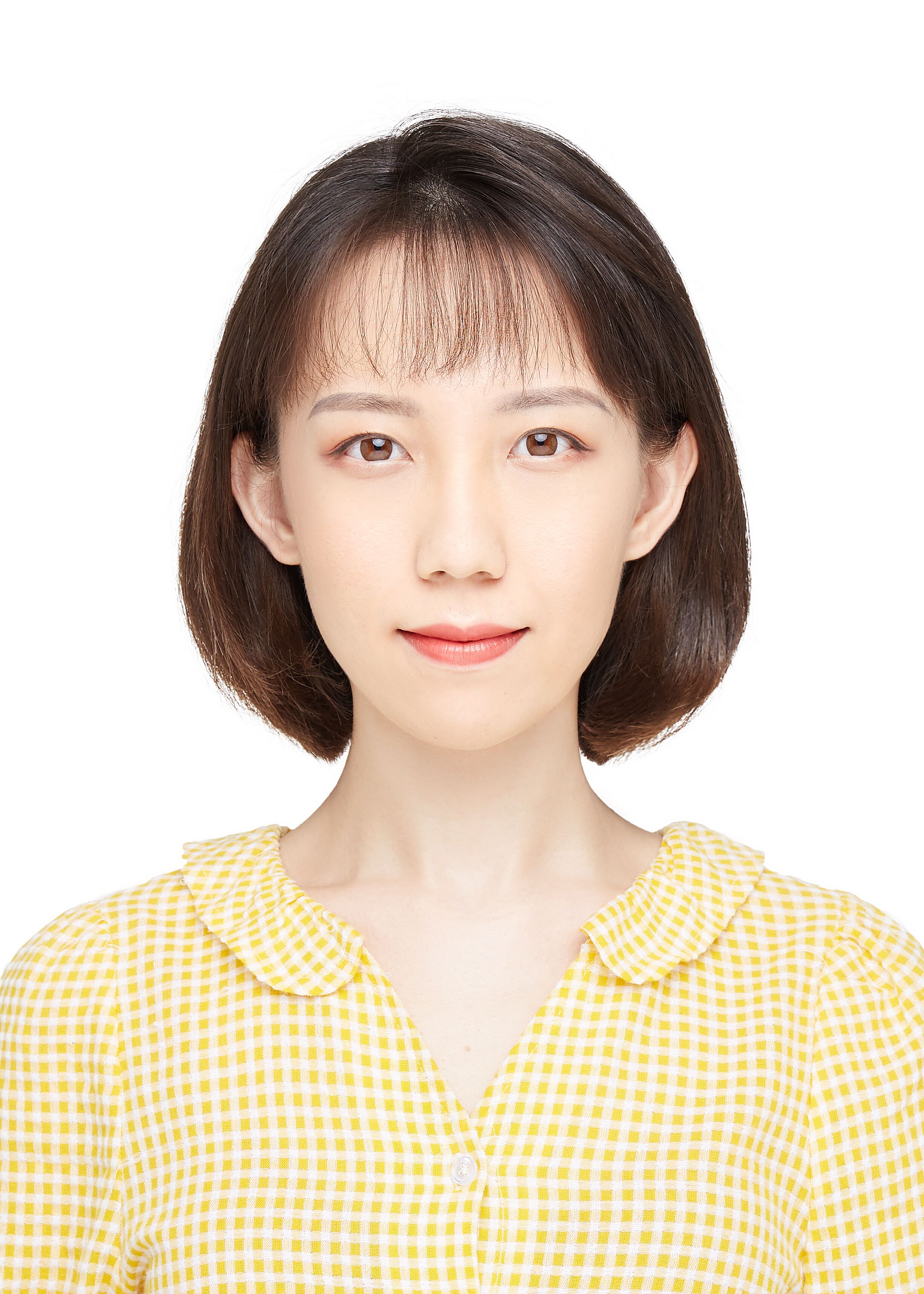}}]{Jingyuan Yang}
	received the B.Eng. degree in Electronic and Information Engineering from Xidian University, Xi'an, China, in 2017. She is currently a Ph. D. Candidate at the School of Electronic Engineering, Xidian University. Her current research interest is visual emotion analysis in deep learning and its applications.
\end{IEEEbiography}

\begin{IEEEbiography}[{\includegraphics[width=1in,height=1.25in,clip,keepaspectratio]{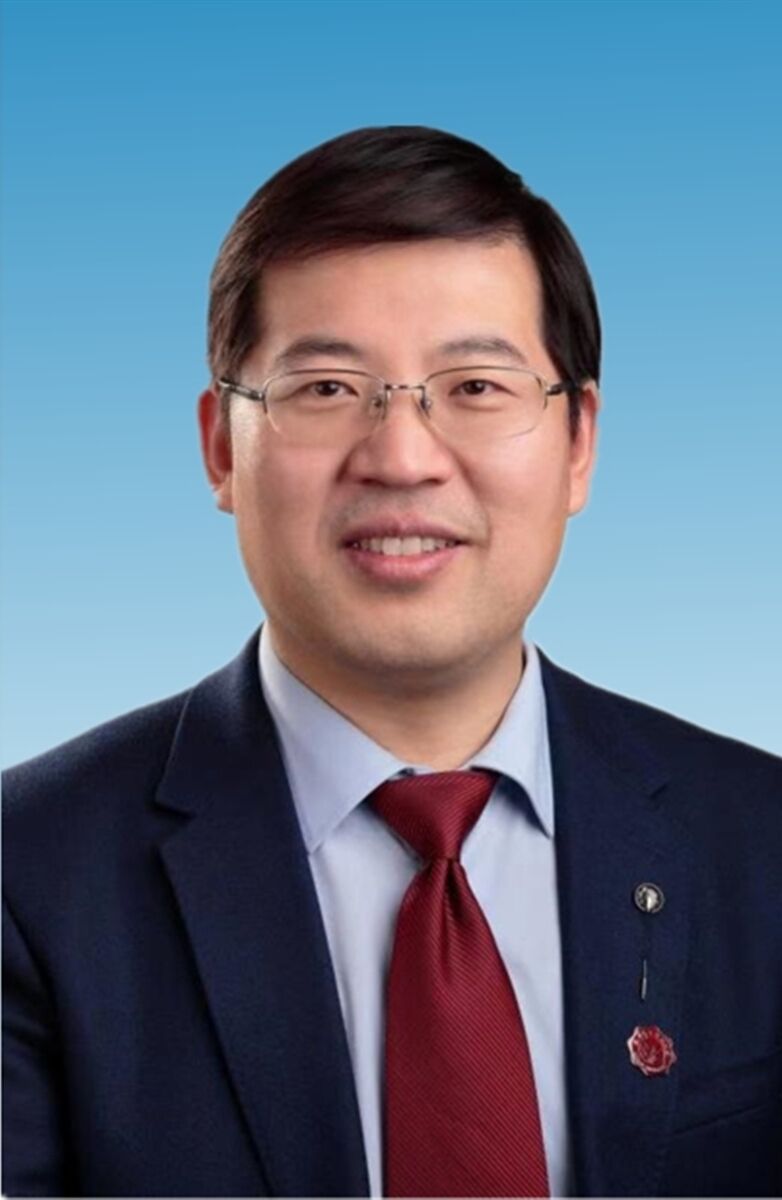}}]{Xinbo Gao}
	(Senior Member, IEEE) received the B.Eng., M.Sc. and Ph.D. degrees in electronic engineering, signal and information processing from Xidian University, Xi'an, China, in 1994, 1997, and 1999, respectively. From 1997 to 1998, he was a research fellow at the Department of Computer Science, Shizuoka University, Shizuoka, Japan. From 2000 to 2001, he was a post-doctoral research fellow at the Department of Information Engineering, the Chinese University of Hong Kong, Hong Kong. Since 2001, he has been at the School of Electronic Engineering, Xidian University. He is a Cheung Kong Professor of Ministry of Education of P. R. China, a Professor of Pattern Recognition and Intelligent System of Xidian University. Since 2020, he is also a Professor of Computer Science and Technology of Chongqing University of Posts and Telecommunications. His current research interests include Image processing, computer vision, multimedia analysis, machine learning and pattern recognition. He has published six books and around 300 technical articles in refereed journals and proceedings. Prof. Gao is on the Editorial Boards of several journals, including Signal Processing (Elsevier) and Neurocomputing (Elsevier). He served as the General Chair/Co-Chair, Program Committee Chair/Co-Chair, or PC Member for around 30 major international conferences. He is a Fellow of the Institute of Engineering and Technology and a Fellow of the Chinese Institute of Electronics.
\end{IEEEbiography}

\begin{IEEEbiography}[{\includegraphics[width=1in,height=1.25in,clip,keepaspectratio]{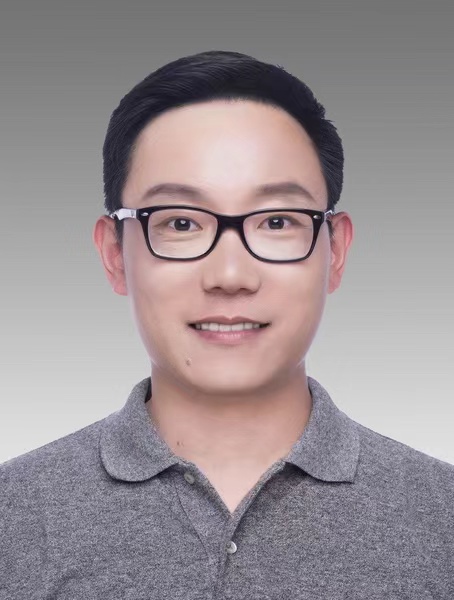}}]{Leida Li}
	(Member, IEEE) received the B.S. and Ph.D. degrees from Xidian University, Xi’an, China, in 2004 and 2009, respectively. In 2008, he was a Research Assistant with the Department of Electronic Engineering, National Kaohsiung University of Science and Technology, Kaohsiung, Taiwan. From 2014 to 2015, he was a Visiting Research Fellow with the Rapid-Rich Object Search (ROSE) Lab, School of Electrical and Electronic Engineering, Nanyang Technological University, Singapore, where he was a Senior Research Fellow, from 2016 to 2017. He is currently a Professor with the School of Artificial Intelligence, Xidian University, China. His research interests include multimedia quality assessment, affective computing, information hiding, and image forensics.
	He has served as an SPC for IJCAI 2019-2021, the Session Chair for ICMR 2019 and PCM 2015, and TPC for CVPR 2021, ICCV 2021, AAAI 2019-2021, ACM MM 2019-2020, ACM MM-Asia 2019, and ACII 2019.
	He is currently an Associate Editor of the Journal of Visual Communication and Image Representation and the EURASIP Journal on Image and Video Processing.
\end{IEEEbiography}

\begin{IEEEbiography}[{\includegraphics[width=1in,height=1.25in,clip,keepaspectratio]{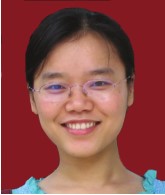}}]{Xiumei Wang}
	received the Ph.D. degree from Xidian University, Xi’an, China, in 2010. She is currently a Lecturer with the School of Electronic Engineering, Xidian University. Her current research interests include nonparametric statistical models and machine learning. She has published several scientific articles, including the IEEE Trans. Cybernetics, Pattern Recognition, and Neurocomputing in the above areas.
\end{IEEEbiography}

\begin{IEEEbiography}[{\includegraphics[width=1in,height=1.25in,clip,keepaspectratio]{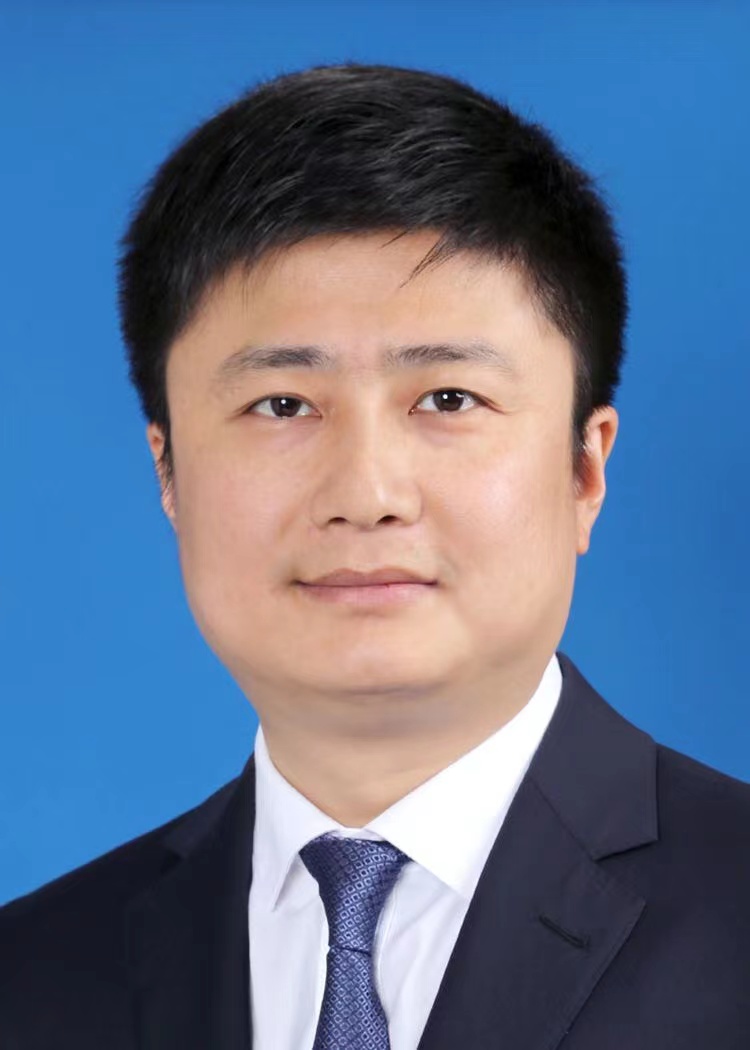}}]{Jinshan Ding}
	is a Professor with the School of Electronic Engineering, Xidian University, Xi’an, China. He founded the Millimeter-Wave and THz Research Group, Xidian University, after his return from Germany. His research interests include millimeter-wave and THz radar, video SAR, and machine learning in radar.
\end{IEEEbiography}

% that's all folks
\end{document}